\RequirePackage{fix-cm}
\documentclass[twocolumn]{svjour3}          % twocolumn
\smartqed  % flush right qed marks, e.g. at end of proof
\usepackage{graphicx}

\usepackage{amsmath}
\usepackage{amssymb}
\usepackage{booktabs}
\usepackage[pagebackref=true,breaklinks=true,letterpaper=true,colorlinks,bookmarks=false]{hyperref}
\usepackage{multirow}
\usepackage{cite}
\DeclareUnicodeCharacter{211D}{\mathbb{R}}

\newcommand{\etal}{~\textit{et al.}}

\newcommand{\eg}{\textit{e.g}}

\journalname{IJCV}
\begin{document}

\title{CRCNet: Few-shot Segmentation with Cross-Reference and Region-Global Conditional Networks
}

\author{Weide~Liu \and
        Chi Zhang \and
        Guosheng~Lin \and
        Fayao Liu
}

%\authorrunning{Short form of author list} % if too long for running head

\institute{W. Liu is with School of Computer Science and Engineering, Nanyang Technological University (NTU), Singapore 639798 and with Institute for Infocomm Research, A*STAR, Singapore 138632 (e-mail: weide001@e.ntu.edu.sg).         \\  %  \\
%             \emph{Present address:} of F. Author  %  if needed
        %   \and
           C. Zhang is with School of Computer Science and Engineering, Nanyang Technological University (NTU), Singapore 639798 (e-mail: chi007@e.ntu.edu.sg). \\
        %   \and
           G. Lin is with School of Computer Science and Engineering, Nanyang Technological University (NTU), Singapore 639798 (e-mail: gslin@ntu.edu.sg). \\
        %   \and
           F. Liu is with Institute for Infocomm Research, A*STAR, Singapore 138632
(e-mail: liu\_fayao@i2r.a-star.edu.sg). \\
Corresponding author: Guosheng Lin.
}

\date{Received: date / Accepted: date}

\maketitle

\begin{abstract}
Few-shot segmentation aims to learn a segmentation model that can be generalized to novel classes with only a few training images. 
In this paper, we propose a Cross-Reference and Local-Global Conditional Networks (CRCNet) for few-shot segmentation. Unlike previous works that only predict the query image's mask, our proposed model concurrently makes predictions for both the support image and the query image. Our network can better find the co-occurrent objects in the two images with a cross-reference mechanism, thus helping the few-shot segmentation task. 
To further improve feature comparison, we develop a local-global conditional module to capture both global and local relations. We also develop a mask refinement module to refine the prediction of the foreground regions recurrently. Experiments on the PASCAL VOC 2012, MS COCO, and FSS-1000 datasets show that our network achieves new state-of-the-art performance.
\end{abstract}
\section{Introduction}
Deep neural networks have been widely applied to visual understanding tasks, \eg, image classification~\cite{he2016deep,krizhevsky2012imagenet}, objection detection~\cite{ren2015faster}, semantic segmentation~\cite{lin2017refinenet} and image captioning, since the huge success in ImageNet classification challenge~\cite{imagenet}. Due to its data-driving property, large-scale labeled datasets are often required to train deep models. 
However, collecting labeled data can be notoriously expensive in tasks with pixel-level annotations like semantic segmentation, instance segmentation, and video segmentation. Moreover, data collecting is usually for a set of specific categories. Knowledge learned in previous classes can hardly be transferred to unseen classes directly.
Directly finetuning the trained models still needs a large amount of new labeled data. Few-shot learning, on the other hand, is proposed to solve this problem.
In this paper, we target at few-shot image segmentation. Few-shot segmentation aims to find the foreground regions of the category the same as the provided support image, only seeing a few labeled examples.

 \begin{figure}[t]
  \centering
    \includegraphics[width=1\linewidth]{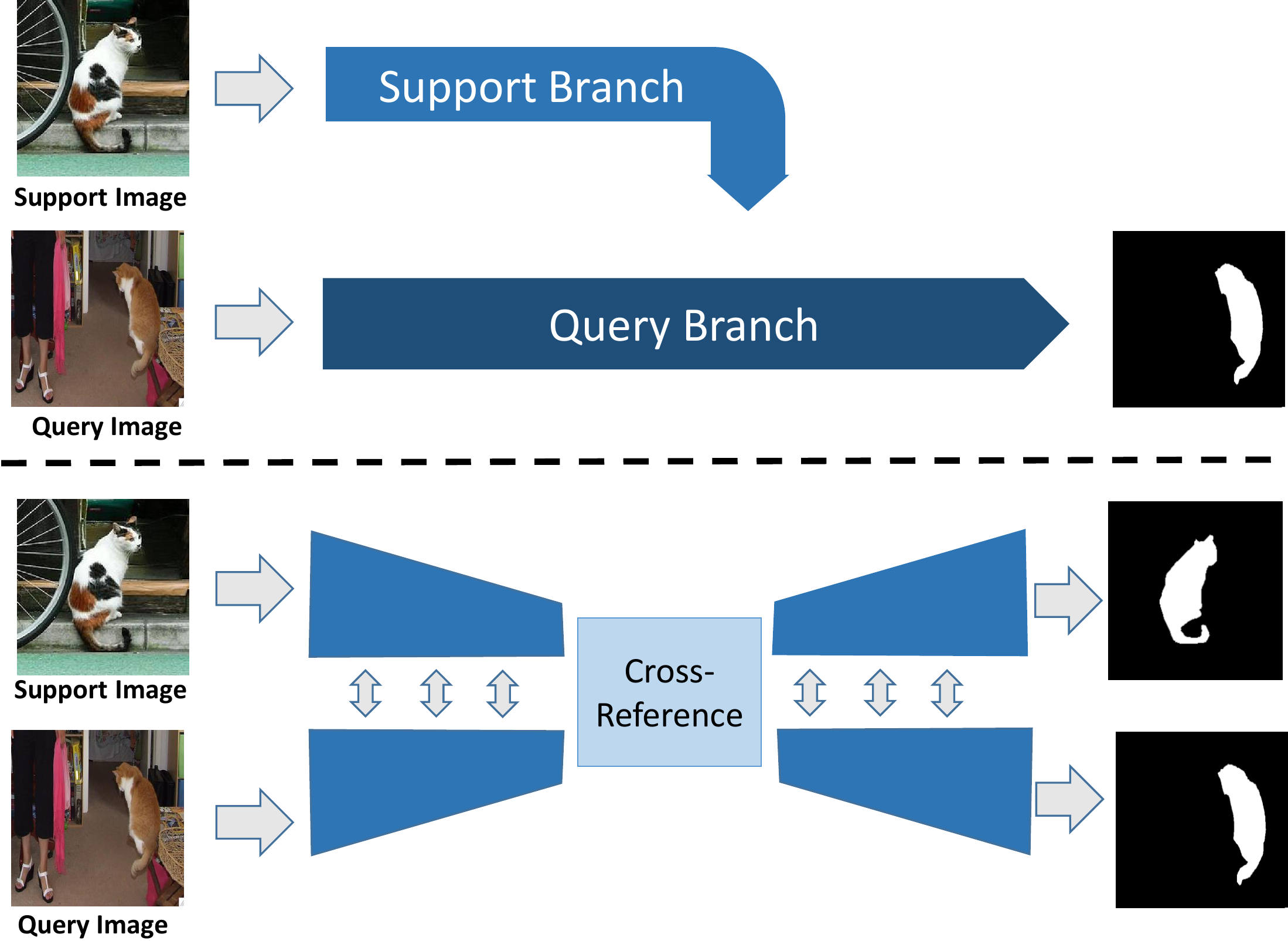}
    \caption{Comparison of our proposed CRCNet against previous work. Previous works (upper part) unilaterally guide the segmentation of query images with support images, while in our CRCNet (lower part) support and query images can guide the segmentation of each other. }
    \label{twopair}
\end{figure}

Many previous works~\cite{zhang2019canet,Dong2018FewShotSS,zhangchi2,fss1000} formulate the few-shot segmentation task as a guided segmentation task. The guidance information is extracted from the labeled support set for the foreground prediction in the query image, usually achieved by an unsymmetrical two-branch network structure. The model is optimized with the ground truth query mask as the supervision. However, such an unsymmetrical structure only leverages the support image to guide the query image, which does not fully use the support image and query image information. Our work argues that the roles of query and support sets can be switched in few-shot segmentation models. Specifically, the support images can guide the query set's prediction, and conversely, the query image can also help make predictions of the support set. Inspired by the image co-segmentation literature~\cite{joulin2012multi,mukherjee2018object,chen2018semantic,tip-co-seg1,tip-co-seg2}, we propose a symmetric Cross-Reference Network that two heads concurrently make predictions for both the query image and the support image.
The difference of the network design with previous works is shown in Figure~\ref{twopair}.
Our network design's critical component is the cross-reference module, which generates the reinforced feature representations by comparing the two images' co-occurrent features. The reinforced representations are used for the downstream foreground predictions in two images. In the meantime, the cross-reference module also makes predictions of co-occurrent objects in the two images.  This sub-task provides an auxiliary loss in the training phase to facilitate the training of the cross-reference module. 

Metric learning has been widely used in the few-shot learning~\cite{zhang2019canet,snell2017prototypical}; however, it is an enormous time cost to make comparisons between all pixels pairs. To alleviate this problem, we capture the global representation from the support image set with foreground average pooling operation to retrieve the category information in query images. In particular, we fuse the feature of the target category to generate the category-relevant vector. As the few-shot segmentation's goal is only to find the foreground mask of the assigned object category, the task-relevant vector serves as a condition to segment the target category. 

Only incorporating the global representation from the support image with the query image will cause several limitations. It will ignore the local similarity between object parts. Due to occlusions, viewpoint variations, and other reasons, one object may be only partially similar to other objects of the same category  -- only some parts are similar. Global features may not be able to capture such part-level similarity relations.
To alleviate these limitations, we investigate the region-level comparison between query and support images to further enhance the conditional module. 
Inspired by~\cite{zhangchi2}, we develop a local-global conditional module to efficiently incorporate the global information and local information for the comparison module. 

As there is a massive variance in the object appearance, mining foreground regions in images can be multi-step. We develop an effective Mask Refinement Module to refine our predictions iteratively. 
In the initial prediction, the network is expected to locate high-confidence seed regions. The confidence map is then saved as the cache in the module and is used for later predictions. We update the cache every time we make a new prediction. Our model can better predict the foreground regions after running the mask refinement module for a few steps. We empirically demonstrate that such a lightweight module can significantly improve performance.

When it comes to the $k$-shot image segmentation where more than one support image is provided, previous methods~\cite{zhang2019canet,fss1000,Dong2018FewShotSS} often use a 1-shot model to make predictions with each support image individually and fuse their features or predicted masks. 
In our paper, we propose to finetune parts of our network with the labeled support examples. As the network can make predictions for both image inputs, we can use at most $k^2$ image pairs to finetune our network. An advantage of our finetuning-based method is that it can benefit from increasing support images and consistently increasing accuracy. In comparison, fusion-based methods can quickly saturate when more support images are provided. In our experiment, we validate our model in the 1-shot, 5-shot, and 10-shot settings.

The main contributions of this paper are listed as follows: 
\begin{itemize}
    \item We propose a novel cross-reference network that concurrently makes predictions for both the query set and the support set in the few-shot image segmentation task. By mining the co-occurrent features in two images, our proposed network can effectively improve the results.
\item We develop a mask refinement module with a confidence cache that is able to refine the predicted results recurrently.
\item We propose a finetuning scheme for $k$-shot learning, which turns out to be an effective solution to handle multiple support images. 
\item We develop a local-global conditional module that captures both local and global relations between query and support images to gain further improvement. 
\end{itemize}

Parts of the results in this paper were published in its conference version~\cite{crnet}. This paper extends our earlier work in several important aspects:

\begin{itemize}
    \item We extend the proposed CRNet~\cite{crnet} algorithm with a local-global conditional module to efficiently incorporate the global and local information for the query image mask predictions. 
    \item We conduct a new experiment on FSS-1000 (Table~\ref{fss-results}) and MS COCO (Table~\ref{table:coco-bfiou}, Table~\ref{Table:coco-soa-new}, and Table~\ref{coco-soa}) to further evaluate our algorithm.
    \item We conduct a new experiment on a few shot weakly-supervised segmentation tasks. The experiments show that our new algorithms achieve a new state of the art on both Pascal VOC and FSS-1000 datasets (Table~\ref{weakly-1-shot}).
    \item We present comprehensive ablation studies and analysis, such as analysis of the effectiveness of Sigmoid in cross-reference module (Table~\ref{Table: abalation-sigmoid}), the experiments on how the effectiveness of support mask in the local conditional module (Table~\ref{Table: abalation-local-condition-mask}), the finetuning scheme effectiveness to PFENet (Table~\ref{Table:PFE-FT}), more ablation studies on Pascal VOC (Table~\ref{Table:voc-soa-miou}) and MS COCO (Table~\ref{Table:coco-soa-new}) to investigate each component of our network.
\end{itemize}
\section{Related Work}
\subsection{Few shot learning}
Few-shot learning aims to learn a model that can be easily transferred to new tasks with limited training data available. Few-shot learning is widely explored in image classification tasks. Previous methods can be roughly divided into two categories based on whether the model needs finetuning at the testing time. In non-finetuned methods, parameters learned at the training time are kept fixed at the testing stage. For example, ~\cite{snell2017prototypical,relation,vinyals2016matching} are metric-based approaches with an embedding encoder to learn a distance metric to determine the image pair similarity. These methods have the advantage of fast inference without further parameter adaptions. However, when multiple support images are available, the performance can become saturate easily. In finetuning-based methods, the model parameters need to be adapted to the new tasks for predictions. For example,  in~\cite{chen2019closerfewshot}, they demonstrate that by only finetuning the fully connected layer, models learned in training classes can yield state-of-the-art few-shot performance in new classes. 
The usage of attention mechanisms is widespread in few-shot classification and detection tasks. For example, CAN~\cite{review3_2} and \cite{review3_4} exploit the similarity between the few shot support set and a query set among the spatial direction between the images with the spatial attention, while our CRCNet aims to mine out the co-occurrent objects between the images among the channel space with the channel attention.

\subsection{Segmentation}
Semantic segmentation is a fundamental computer vision task that aims to classify each pixel in the image. State-of-the-art methods formulate image segmentation as a dense prediction task and adopt fully convolutional networks to make predictions~\cite{long2015fully}. Usually, a pre-trained classification network is used as the network backbone by removing the fully connected layers at the end. To make pixel-level dense predictions, encoder-decoder structures~\cite{lin2017refinenet,long2015fully,refine-tpami,pfenet,liu2020guided,liu2020weakly,liu2021cross,liu2021few,liu2022long,hou2022distilling,hou2022interaction,zhang2020splitting,zhang2021meta,zhang2020deepemd} are often used to reconstruct high-resolution prediction maps. Typically, an encoder gradually downsamples the feature maps, acquiring a large field-of-view and capturing abstract feature representations. Then, the decoder gradually recovers the fine-grained information. Skip connections are often used to fuse high-level and low-level features for better predictions. In our network, we also follow the encoder-decoder design, transfer the guidance information in the low-resolution maps, and use decoders to recover details.
Global and local information has been widely used in segmentation and detection tasks. The GLNet~\cite{review3_1} effectively preserves both global and local information to capture the high-resolution structures and the contextual information to handle the ultra-high resolution image segmentation tasks. While the MEGA~\cite{review3_3} also takes both global and local information with a long-range memory module. In contrast to those methods, our CRCNet aims to capture the global and local information between different images (the information between the support and query images in the few shot segmentation tasks). 

\subsection{Few-shot segmentation}
Few-shot segmentation~\cite{citeb,citec,cited} is a natural extension of a few-shot classification to pixel levels. Since Shaban~\etal~\cite{shaban2017one} proposes this task for the first time, many deep learning-based methods are proposed. Most previous works formulate the few-shot segmentation as a guided segmentation task. For example, in~\cite{shaban2017one},  the side branch takes the labeled support image as the input and regresses the network parameters in the main branch to make foreground predictions for the query image. In~\cite{zhang2019canet}, they share the same spirits and propose to fuse the embeddings of the support branches into the query branch with a dense comparison module.
Dong~\etal~\cite{Dong2018FewShotSS} draws inspiration from the success of Prototypical Network~\cite{snell2017prototypical} in few-shot classification and proposes a dense prototype learning with Euclidean distance as the metric for segmentation tasks. Similarly, Zhang~\etal~\cite{zhang2018sg} proposes a cosine similarity guidance network to weight features for the foreground predictions in the query branch. 
QGNet~\cite{QGNet} aims to extract the information from the query itself independently with global-local contrastive learning to benefit the few-shot segmentation task. A global contrastive loss is applied to the global representation (image features) to minimize the distance of features obtained by different variants of an identical image. To obtain the local features, the QGNet~\cite{QGNet} applies a local contrastive loss between the local patches to maximize the local feature distance between different patches. Unlike QGNet, in this paper, we obtain the global features with a masked average pooling on the support images and aim to obtain the local features from the region-level comparison between query and support images.
Some previous works use recurrent structures to refine the segmentation predictions~\cite{hu2019attention,zhang2019canet}. We also adapt the mask refinement mechanism, which applies a cache module to optimize the predictions by attending to the confidence maps in the cache. The motivation is that a single feed-forward prediction can reflect the confident region, which serves as the seed region and is saved in the cache. We apply Encoder-Decoder architecture to combine the cached confident regions and features output from previous modules. To better fuse the seeding regions and feature maps, we should let them interact in multiple feature levels and field-of-views. Therefore, skip connections and multi-level features are utilized for better fusion. While CANet~\cite{zhang2019canet} simply concatenate the predicted mask into the intermediate layers.
PANet~\cite{wang2019panet} aims to generate prototypes to perform segmentation through pixel-level matching within the embedding space. The prototypes are obtained by embedding different foreground objects and backgrounds into different prototypes via a shared feature extractor. In contrast, our method aims to mine out the co-occurrent objects through the cross-reference model and the conditional module. In addition, the PANet also finds that the features generated from the query sets also benefit the support sets.

Most of the previous methods only use the foreground mask in the query image as the training supervision. In contrast, in our network, the query set and the support set guide each other, and both branches make foreground predictions for training supervision.

\subsection{Image co-segmentation}
Image co-segmentation is a well-studied task that aims to segment the common objects in paired images jointly. Many approaches have been proposed to solve the object co-segmentation problem.  Rotheret~\etal~\cite{rother2006cosegmentation} proposes to minimize the energy function of a histogram matching term with an MRF to enforce similar foreground statistics. Rubinsteinet~\etal~\cite{rubinstein2013unsupervised} captures the sparsity and observed variability of the common object from pairs of images with dense correspondences. Joulin ~\etal~\cite{joulin2012multi} solves the common object problem with an efficient convex quadratic approximation of energy with discriminate clustering. 
Since the prevalence of deep neural networks, many deep learning-based methods have been proposed. In~\cite{mukherjee2018object}, the model retrieves common object proposals with a Siamese network. Chen~\etal~ \cite{chen2018semantic} adopt channel attention to weight features for the co-segmentation task. Deep learning-based approaches have significantly outperformed non-learning-based methods. In this work, we argue that the few shot segmentation tasks can be considered similar to finding the co-occurred object between query and support image sets.
\section{Task Definition} \label{problem_define}
Few-shot segmentation aims to find the foreground pixels in the test images given only a few pixel-level annotated images. The training and testing of the model are conducted on two datasets with no overlapped categories. At both the training and testing stages, the labeled example images are called the support set, which serves as a meta-training set, and the unlabeled meta-testing image is called the query set. 

Given a network $\mathcal{R}_{\theta}$ parameterized by ${\theta}$, in each episode, we first sample a target category $c$ from the dataset $\mathcal{C}$. Based on the sampled class, we then sample $k+1$ labeled images $\{(x_s^1,y_s^1),(x_s^2,y_s^2),...(x_s^k,y_s^k),(x_q,y_q) \}$ that all contain the sampled category $c$. Among them, the first $k$ labeled images constitute the support set $\mathcal{S}$ and the last one is the query set $\mathcal{Q}$. After that, we make predictions on the query images by inputting the support set and the query image into the model $\hat y_{q}=\mathcal{R}_{\theta}(\mathcal{S},x_q)$. At training time, we learn the model parameters $\theta$ by optimizing the cross-entropy loss $\mathcal{L}(\hat y_{q},y_{q})$, and repeat such procedures until convergence.
\section{Method}
 \begin{figure*}[t]
 \centering
    \includegraphics[width=1\linewidth]{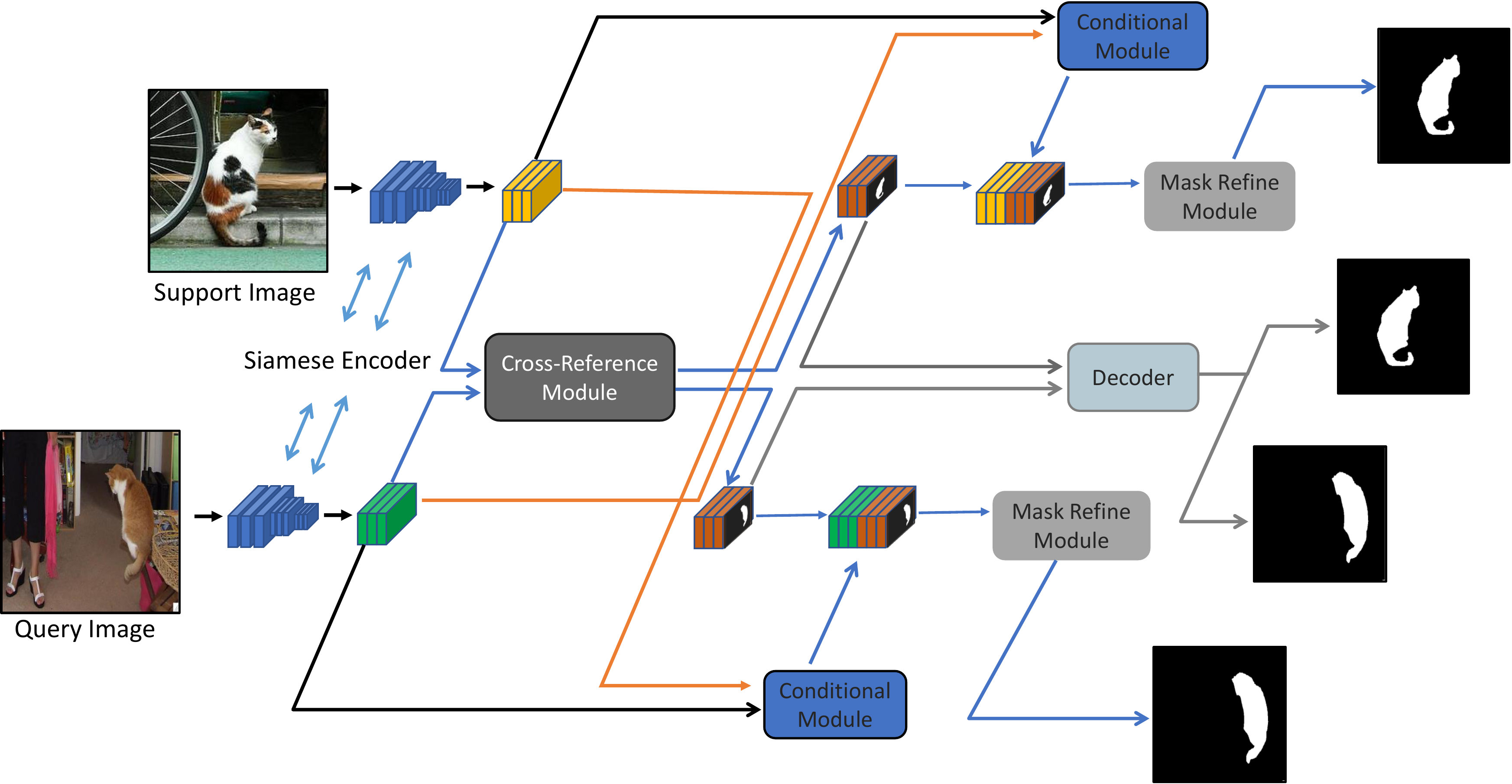}
    \caption{The pipeline of our network architecture. Our network mainly consists of a Siamese Encoder, a cross-reference module, a conditional module, and a mask refinement module. Our network adopts a symmetric design. 
The Siamese encoder maps the query and support images into feature representations. 
The cross-reference module mines the co-occurrent features in two images to generate reinforced representations. 
The conditional module fuses the category-relevant feature vectors into feature maps to emphasize the target category. The mask refinement module saves the confidence maps of the last prediction into the cache and recurrently refines the predicted masks.}
    \label{archtecture}
\end{figure*}
In this section, we will introduce the proposed cross-reference and local-global conditional network for solving few-shot image segmentation. 
In the beginning, we describe our network in the 1-shot case. 
After that, we describe our finetuning scheme in the case of $k$-shot learning. Our network includes four key modules: the Siamese encoder, the cross-reference module, the conditional module, and the mask refinement module. The overall architecture is shown in Figure~\ref{archtecture}.

\subsection{Method overview}

Unlike previous existing few-shot segmentation methods \cite{zhang2019canet,shaban2017one,Dong2018FewShotSS,mdl,fss1000} unilaterally guide the segmentation of query images with support images, our proposed CRCNet enables support, and query images guide the segmentation of each other. 
We argue that the relationship between support-query image pairs is vital to few-shot segmentation learning. Experiments in Table~\ref{Table:ablation-condition-cr} validate the effectiveness of our new architecture design.
As shown in Figure~\ref{archtecture}, for every query-support pair, we encode the image pair into the features with the Siamese encoder, then apply the cross-reference module to mine out co-occurrent objects features. To fully utilize the annotated mask, the conditional module will incorporate the category information of support set annotations for foreground mask predictions. Finally, our mask refines module caches the confidence region maps recurrently for final foreground prediction. 
In the case of $k$-shot learning, previous works \cite{zhang2018sg,zhang2019canet,shaban2017one} only simply average the results of different 1-shot predictions. In contrast, our CRCNet adopts an optimization-based method that finetunes the model to use more support data. Table~\ref{ablation:Fuse-and-FT} demonstrates the advantages of our method over previous works.

\subsection{Siamese encoder}
The Siamese encoder is a pair of parameter-shared convolutional neural networks that encode the query and support images to feature maps. Unlike the models in~\cite{shaban2017one,rakelly2018conditional}, we use a shared feature encoder to encode the support and the query images. Our cross-reference module can provide better co-occurrent mine features to locate the foreground regions by embedding the images into the same space. To acquire representative feature embeddings, we use skip-connections to utilize multiple-layer features. As observed in CNN feature visualization literature\cite{zhang2019canet,yosinski2015understanding}, features in lower layers often relate to low-level cues, and higher layers often relate to segment cues combined the lower level features and higher-level features and passing to followed modules.

\subsection{Cross-Reference Module} \label{sec:cross-ref}
\begin{figure}[t]
  \centering
    \includegraphics[width=1\linewidth]{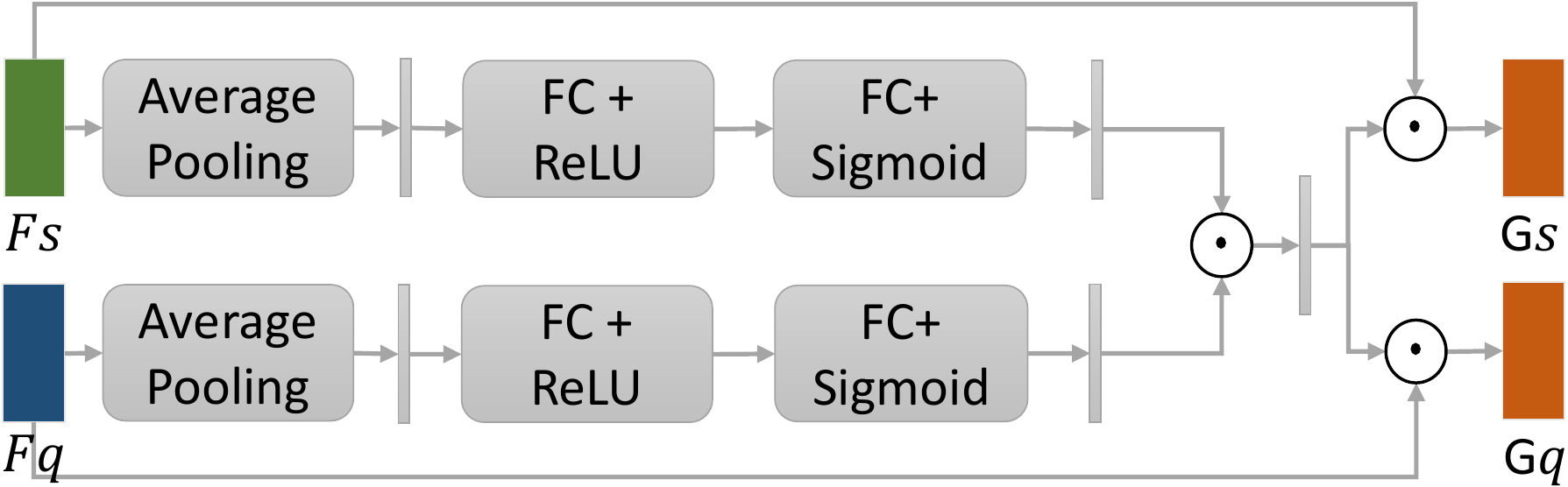}
    \caption{The cross-reference module. Given the input feature maps from the support and the query sets ($F_s$,$F_q$), the cross-reference module generates updated feature representations ($G_s$,$G_q$) by inspecting the co-occurrent features.}
    \label{co-segmentation}
\end{figure}

The cross-reference module is designed to mine co-occurrent features in two images and generate updated representations. The design of the module is shown in Figure~\ref{co-segmentation}. Given two input feature maps generated by the Siamese encoder, we first use global average pooling to acquire the two images' global statistics. The two feature vectors are then sent to two fully connected (FC) layers, respectively. The Sigmoid activation function attached after the last FC layer transforms the vector values into the channel's importance, which is in the range of [0,1]. After that, the vectors in the two branches are fused by element-wise multiplication. Intuitively, only the two branches' common features will highly activate the fused importance vector. Finally, we use the fused vector to re-weight the input feature maps to generate reinforced feature representations. In comparison to the basic features, the reinforced features focus more on the co-occurrent representations.  

We add a head to directly predict the two images' co-occurrent objects during training time based on the reinforced feature representations.  This sub-task aims to facilitate the co-segmentation module's learning to mine better feature representations for the downstream tasks. To generate the co-occurrent objects' predictions in two images, the reinforced feature maps in the two branches are sent to a decoder to generate the predicted maps $QM_{sub}$ and $SM_{sub}$. 
The decoder is composed of several convolutional layers and ASPP~\cite{chen2018deeplab} layers. Finally, we generate a two-channel prediction with a convolutional layer corresponding to the foreground and background scores. 

\subsection{Conditional Module}
\begin{figure}[t]
  \centering
%   \vskip +1em
    \includegraphics[width=1\linewidth]{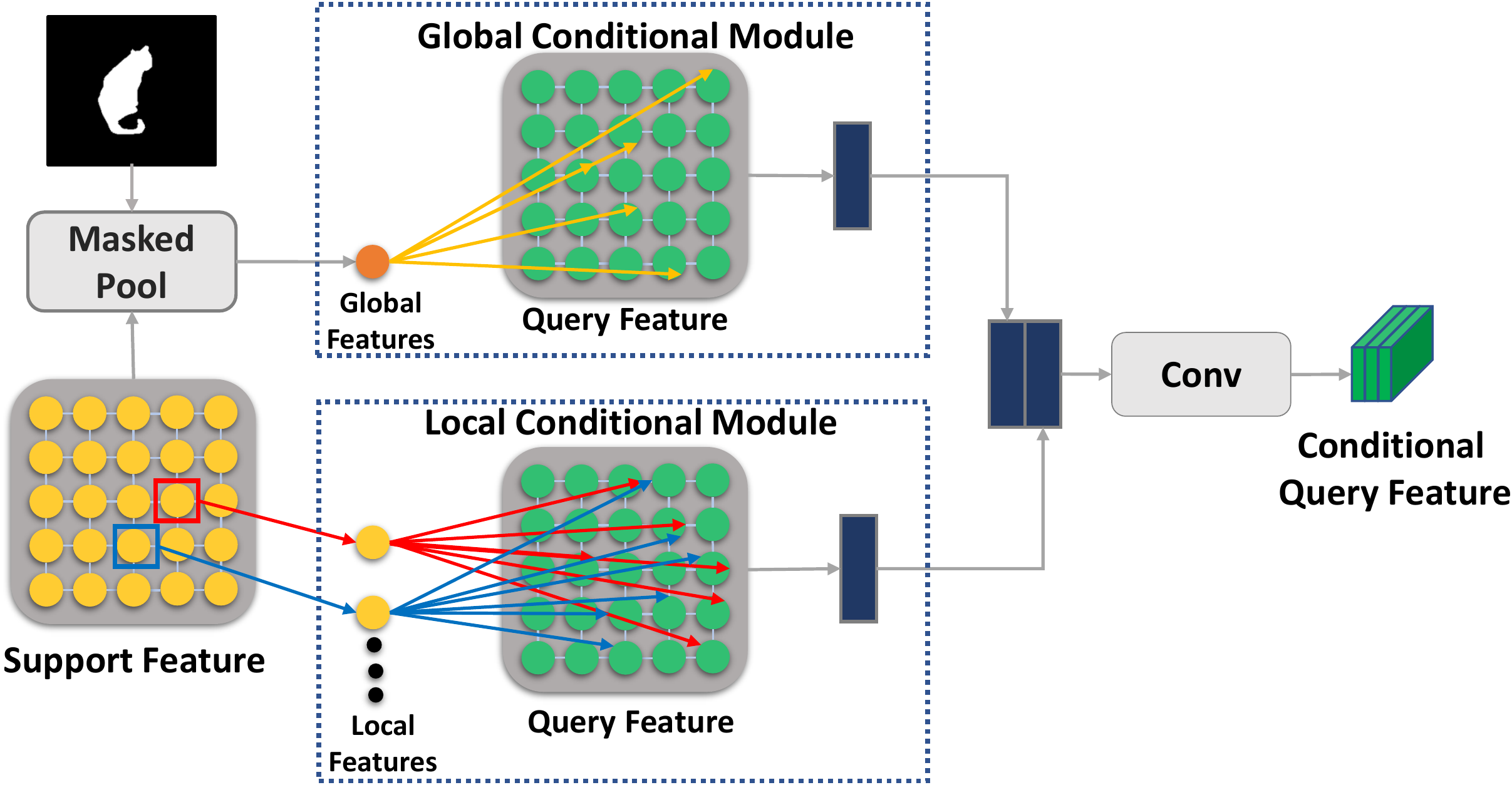}
    \caption{Illustration of the conditional module. Given a pair of support and query features, we first generate the global vector with a masked global average pooling as a global condition to guide the query image prediction. After that, we leverage the spatial element-wise support representations as a local condition to capture local similarity. Finally, we fuse the global and local conditional features to improve feature comparison.}
    \label{Figure:condition}
\end{figure}

As shown in Figure~\ref{Figure:condition}, we design a conditional module to incorporate the category information for foreground mask predictions efficiently. Our Conditional module is composed of global and local conditional modules. Given a support and query image pair, we first use a Siamese encoder to extract their features. Then we use the support annotations to filter out the irrelevant support features. We generate the global category-relevant vector with a masked global average pooling as a global condition to guide the query prediction. After that, we leverage the spatial support features as a local condition to enhance feature comparison. Finally, we fuse the global and local conditional features into representations to better predict the query masks. 

\begin{figure*}[t]
  \centering
    \includegraphics[width=0.8\linewidth]{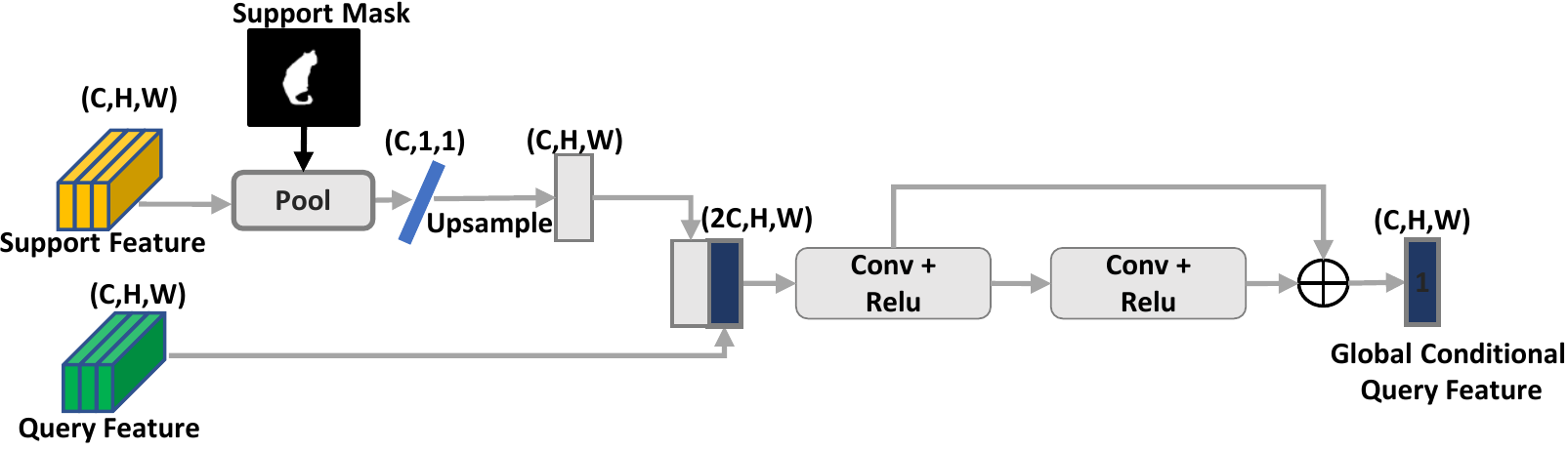}
    \caption{Illustration of the global conditional module. We generate the category-relevant vector by applying foreground average pooling over the category region. Our global conditional module takes the vector as a global condition to guide the query prediction. }
    \label{global-condition}
\end{figure*}

\textbf{Global conditional module.}
The global conditional module takes the feature representations generated by the Siamese encoder and a category-relevant vector as inputs. The category-relevant vector is the fused feature embedding of the target category, which is achieved by applying foreground average pooling~\cite{zhang2019canet} over the category region. As the few-shot segmentation's goal is only to find the foreground mask of the assigned object category, the task-relevant vector serves as a condition to segment the target category.
The structure of our global conditional module is shown in Figure~\ref{global-condition}.  At first, apply a bilinearly upsampling to the category-relevant vector to the feature maps' same spatial size. Then we concatenate them with the query features. After that, we use a residual convolution to process the concatenated features.  The global conditional modules in the support branch and the query branch have the same structure.

\begin{figure*}[t]
  \centering
    \includegraphics[width=0.9\linewidth]{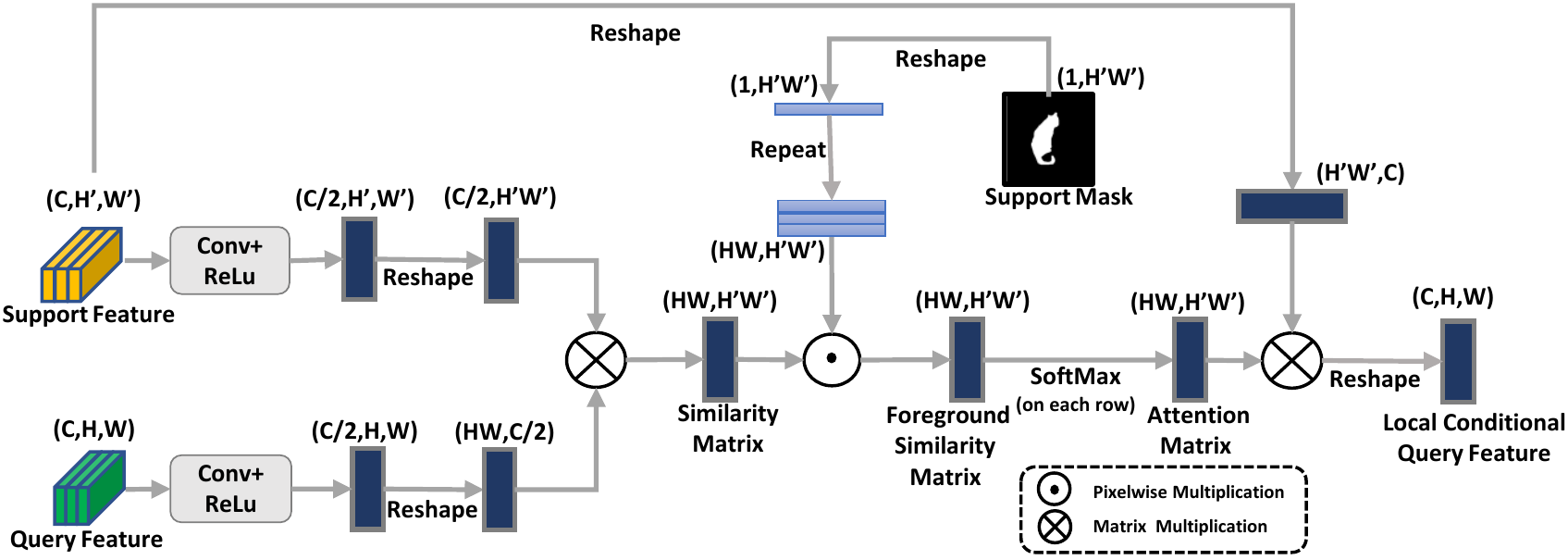}
    \caption{Our local conditional module leverages the spatial features as a condition to capture local similarities. Given a pair of support and query features, we first encoder them with a $1\times1$ convolution layer, then we generate the similarity matrix with a matrix multiplication operation. The row of the similarity matrix indicates the similarity of each query feature to the support features of all spatial positions. With the support mask, we generate the foreground similarity matrix. After that, we apply a Softmax on each row of the similarity matrix to generate the attention matrix. Combined with the support feature, we generate a similarity feature map to improve the conditional module.}
    \label{local-condition}
\end{figure*}

\textbf{Local conditional module.}
The cross-reference model aims to mine out the co-occurrent objects between the images among the channel space. In particular, to mine the co-occurrent regions of two feature maps, the cross-reference model exchanges global information, generating weights for the channels. To complement the cross-reference module, we propose a local conditional model that aims to leverage the region-level comparison between query and support images to further enhance the co-occurrent objects among the spatial dimensions.
Our local conditional module leverages the spatial element-wise representations as a condition to capture local similarities. As is depicted in Figure~\ref{local-condition}, the local conditional module takes the query and support image features ($F_q$, \textit{resp.}$F_s$) generated by the Siamese encoder as inputs. We generate a similarity matrix $M_{sim}$ between the query and support representations as follows:

\begin{equation}
    M_{sim}=\theta(F_q) ^{T} \bigotimes  \delta (F_s) .
\end{equation}

Here $\theta$ and $\delta$ denote the non-linear transfer operation, which is implemented with 1$\times$ 1 convolutional layer followed by a ReLu activation layer, and $\bigotimes$ denotes the matrix multiplication operation. The row of the similarity matrix indicates the similarity of each query feature to all the support features. 
However, the similarity matrix $M_{sim}$ contains all the support features, including foreground and background, but we only need to enhance the same category features. We filter out the background and generate a foreground similarity matrix by reshaping the support annotation and expanding to the same size as $M_{sim}$, then multiplying to the similar matrix $M_{sim}$. After that, the foreground similarity matrix is normalized row-wise to derive an attention matrix $M^{'}_{sim}$ for each position in the query feature to support features. 

\begin{equation}
    M^{'}_{sim}=Softmax(M_{sim} \cdot \Phi (Mask_{sup}) ).
\end{equation}

Here $\cdot$ denotes the element-wise multiplication operation, and $\Phi (Mask_{sup})$ denotes the reshaped and extended support mask annotation. The softmax is performed row-wise. Finally, we generate the query attention maps $F^{'}_{q}$ with a matrix multiplication to the support feature: 

\begin{equation}
    F^{'}_{q} = F_{s} \bigotimes  M^{'}_{sim}. 
\end{equation}

We concatenate the outputs from the global and local conditional modules to generate our final conditional features. 

\subsection{Mask Refinement Module}
\begin{figure*}[t]
  \centering
    \includegraphics[width=0.8\linewidth]{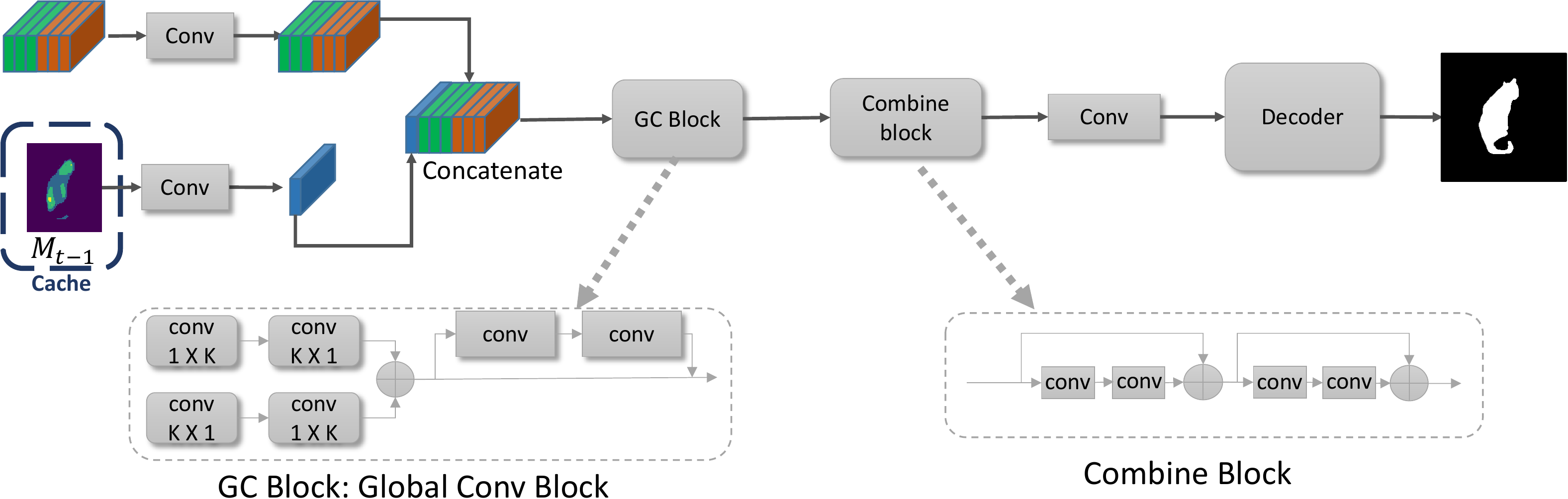}
    \caption{The mask refinement module. The module saves the generated probability map from the last step into the cache and recurrently optimizes the predictions. }
    \label{memory}
\end{figure*}

As is often observed in the weakly supervised semantic segmentation literature~\cite{zhang2019canet,kolesnikov2016seed}, directly predicting the object masks can be difficult. It is a common principle to firstly locate seed regions and then refine the results. Based on such a principle, we design a mask refinement module to refine the predicted mask step-by-step. Our motivation is that the probability maps in a single feed-forward prediction can reflect where is the confident region in the model prediction. We can gradually optimize the mask and find the whole object regions based on the confident regions and the image features. As shown in Figure~\ref{memory}, our mask refinement module has two inputs. One is the saved confidence map in the cache, and the second is the concatenation of the outputs from the conditional and cross-reference modules. The cache is initialized with a zero mask for the initial prediction, and the module makes predictions solely based on the input feature maps. The module cache is updated with the generated probability map every time the module makes a new prediction. We run this module multiple times to generate a final refined mask.

The mask refinement module includes two main blocks: the global convolution block and the combined block.
The global convolution block~\cite{peng2017large} aims to capture features in a large field-of-view while containing few parameters. It includes two groups of $1\times7$ and $7\times1$ convolutional kernels. The combined block effectively fuses the feature branch from different feature levels and the cached branch to generate refined feature representations. 

\subsection{Finetuning for K-Shot Learning}
In the case of $k$-shot learning, we propose to finetune our network to take advantage of multiple labeled support images. As the network can make predictions for two images simultaneously, we can use at most $k^2$ image pairs to finetune our network. We randomly sample an image pair from the labeled support set at the evaluation stage to finetune our model. We kept the parameters in the Siamese encoder fixed and only finetuned the rest modules. Our experiment demonstrates that our finetuning-based methods can consistently improve results when more labeled support images are available. In contrast, previous works' fusion-based methods often get saturated when the number of support images increases.

\subsection{Training}
We choose a pair of images that belong to the same category as support image and the query image during the training process. Their roles are interchangeable. Our training objective includes four Binary Cross-Entropy losses (BCEloss), which are applied on the query predicted masks ( $L_{QM}$ and $L_{QMsub}$) and the support predicted masks ( $L_{SM}$ and $L_{SMsub}$). 
\begin{equation}
BCEloss = - (y*log(p) + (1-y)*(log(1-p)) ) 
\end{equation}
   \begin{equation}
 \mathcal{L}= (\mathcal{L}_{QM}+  \mathcal{L}_{SM}) + \lambda(\mathcal{L}_{QMsub} + \mathcal{L}_{SMsub}).
 \label{loss_sum}
  \end{equation}

Where \textit{p} denotes the predicted probability, and $y$ denotes the ground truth label. $L_{QM}$ and $L_{SM}$ denote the losses with the final predicted query and support masks. $L_{QMsub}$ and $L_{SMsub}$ denote
the losses of the sub-predicted query and support masks discussed in Sec.~\ref{sec:cross-ref}.  We balance the two kinds of losses with hyper-parameter $\lambda$. In this paper, we set $\lambda$ to 0.1.

\begin{table}[]
\centering
\small
\caption{Ablation study on the conditional module and the cross-reference module. The cross-reference module brings a large performance improvement over the baseline model (Conditional module only). In this table, we do not adopt multi-scale, multi-level features and mask refine modules. The results are reported on PASCAL VOC 2012 dataset with standard mIoU(\%). }
\resizebox{0.9\columnwidth}{!}
{

\begin{tabular}{ccccc}
\toprule[1pt]
Conditional Module & Cross-Reference Module  & 1-shot  \\
\hline
\checkmark   &        & 45.8       \\
    & \checkmark      & 46.9         \\
\checkmark   & \checkmark      & 49.6    \\ 
\bottomrule[1pt]
\end{tabular}
}

\label{Table:ablation-condition-cr}
\end{table}

\begin{table}[]
\centering
\small
\caption{Ablation experiments on the multiple-level feature, multiple-scale input, and the mask refine module. Every module brings performance improvement over the baseline model. The results are reported on PASCAL VOC 2012 dataset with standard mIoU(\%).
}
\resizebox{0.9\columnwidth}{!}
{
\begin{tabular}{cccc}
\toprule[1pt]
Multi-Level & Mask Refine & Multi-Scale & 1-shot  \\
\hline
  &    &    & 49.6      \\
  &    & \checkmark  & 50.4       \\
  & \checkmark  & \checkmark  & 53.0     \\
\checkmark  & \checkmark  & \checkmark  & 56.4   \\
\bottomrule[1pt]
\end{tabular}
}
\label{table:abalation-refine-scale}
\end{table}

\begin{table}[]
\centering
\small
\caption{Ablation study on the conditional module. Both global and local conditional module brings a significant performance improvement over the baseline model. The results are reported on PASCAL VOC 2012 dataset with standard mIoU(\%). }
\resizebox{1\columnwidth}{!}
{
\begin{tabular}{ccc}
\toprule[1pt]
Global Conditional Module & Local Conditional Module & 1-shot \\
\hline
         &        & 52.0       \\
\checkmark   &        & 54.7    \\
    & \checkmark      & 55.3        \\
\checkmark   & \checkmark      & 56.4    \\ 
\bottomrule[1pt]
\end{tabular}

}
\label{ablition-condition}
\end{table}

\begin{table}[]
\centering
\small
\caption{$k$-shot experiments. We compare our finetuning-based method with the fusion method. Our method yields consistent performance improvement when the number of support images increases. For the case of $1$-shot, finetune results are not available as CRCNet needs at least two images to apply our finetune scheme. The results are reported on PASCAL VOC 2012 dataset with standard mIoU(\%). 
}
\resizebox{0.85\columnwidth}{!}
{
\begin{tabular}{lccc}
\toprule[1pt]
Method & 1-shot & 5-shot & 10-shot \\
\hline
Fusion   & 49.6   & 50.3   & 49.9    \\
Finetune & N/A   & 54.5   & 58.1  \\
Finetune + Fusion &N/A & 54.6   & 57.8  \\
\bottomrule[1pt]
\end{tabular}
}
\label{ablation:Fuse-and-FT}
\end{table}

\begin{table}[]
\centering
\small
\caption{In this table, we apply the finetuning scheme on the PFENet~\cite{pfenet}. The results are reported on PASCAL VOC 2012 dataset under 5-shot setting with standard mIoU(\%). PFENet denotes that we build our method upon the PFENet~\cite{pfenet}.
FT denotes the finetuning scheme.}
\resizebox{0.45\columnwidth}{!}
{
\begin{tabular}{lc}
\toprule[1pt]
Method & mIoU \\
\hline
PFENet~\cite{pfenet}   &61.9      \\
PFENet + FT    & 62.2         \\
\bottomrule[1pt]
\end{tabular}
}
\label{Table:PFE-FT}
\end{table}

\begin{table}[]
\centering
\small
\caption{Ablation study on the Sigmoid activation function. The results are reported on PASCAL VOC 2012 dataset with standard mIoU(\%) with 1-shot setting. AF denotes the activation function.}
\resizebox{0.35\columnwidth}{!}
{
\begin{tabular}{lc}
\toprule[1pt]
AF & mIoU  \\
\hline
Relu                & 54.6                     \\
Sigmoid             & 56.4                     \\
Softmax             & 55.7    \\ 
\bottomrule[1pt]
\end{tabular}
}
\label{Table: abalation-sigmoid}
\end{table}
\begin{table}[]
\centering
\small
\caption{Ablation study on the effectiveness of the support mask within the local conditional module. The results are reported on PASCAL VOC 2012 dataset with standard mIoU(\%) with 1-shot setting. }
\resizebox{0.37\columnwidth}{!}
{
\begin{tabular}{lc}
\toprule[1pt]
Method & mIoU  \\
\hline
w/ mask                & 56.4                    \\
w/o mask             & 55.8                     \\

\bottomrule[1pt]
\end{tabular}

}

\label{Table: abalation-local-condition-mask}
\end{table}
\begin{table*}[]
\small
\centering
\caption{Detailed 1-shot and 5-shot results on each split under the mIoU(\%) evaluation metric with the PASCAL VOC 2012 dataset. $^*$ denotes the reproduced results. PFENet denotes that we build our method upon the PFENet~\cite{pfenet}. The GC denotes the global conditional module. The CR denotes the cross-reference module, and the CRC denotes utilizing both our conditional module and the cross-reference module. Our method outperforms most state-of-the-art methods and achieves comparable results with the most recent work.
}

\resizebox{\textwidth}{!}{%
\begin{tabular}{l|c|c|ccccc|ccccc}
\toprule[1pt]

\multirow{2}{*}{Method} & \multirow{2}{*}{Training Size} & \multirow{2}{*}{Backbone} & \multicolumn{5}{c|}{1-shot} & \multicolumn{5}{c}{5-shot} \\ \cline{4-13} 
  & & & split-0 & split-1 & split-2 & split-3 & mean & split-0 & split-1 & split-2 & split-3 & mean \\ \hline\hline
OSLSM\cite{shaban2017one} & -- & VGG 16 & 33.6 & 55.3 & 40.9 & 33.5 & 40.8 & 35.9 & 58.1 & 42.7 & 39.1 & 43.9 \\

co-FCN\cite{rakelly2018conditional}      & -- & VGG 16   & 36.7   & 50.6   & 44.9   & 32.4   & 41.1  & 37.5   & 50.0   & 44.1   & 33.9   & 41.4 \\
R-DFCN\cite{siam2019adaptive}  & -- & VGG 16 & 39.2   & 48.0   & 39.2   & 34.2   & 40.2 & 45.3   & 51.4   & 44.9   & 39.5   & 45.3 \\
SG-One\cite{zhang2018sg}   & -- & VGG 16      & 40.2  & 58.4   & 48.4   & 38.4   & 46.3 & 41.9   & 58.6   & 48.6   & 39.4   & 47.1 \\

AMP \cite{amp} & -- &ResNet-Wide & 41.9 & 50.2 & 46.7 & 34.7 & 43.4 & 41.8 & 55.5 & 50.3 & 39.9 & 46.9   \\
FWB\cite{fast-weight} & {512  $\times$ 512}                       & {VGG 16} & 47.0 & 59.6 & 52.6 & 48.3 & 51.9 & 50.9 & 62.9 & 56.5 & 50.1 & 55.1  \\
PANet\cite{wang2019panet}    & {417 $\times$ 417 }                       & {VGG 16}     &42.3 & 58.0   & 51.1   & 41.2   & 48.1 &51.8 & 64.6   & {59.8}   & 46.5   & 55.7 \\

CANet~\cite{zhang2019canet} & {321 $\times$ 321 }    & {Resnet 50}& 52.5 &  {65.9} &  51.3 &  {51.9} &  55.4 &  55.5 &  67.8 &  51.9 &  53.2 &  57.1 \\ 

RPMMs~\cite{pmm} & {321 $\times$ 321 }    & {Resnet 50} & 55.1   & 66.9   & 52.6   & 50.7   & 56.3 & 56.2   & 67.3   & 54.5   & 51.0   & 57.3 \\ 

PPNet~\cite{ppnet} & {417 $\times$ 417 }    & {Resnet 50} & 48.5  & 60.1   & {55.7}   & 46.4   & 52.8 & 58.8   & 68.3   & {66.7}   & 57.9   & 62.9 \\ 

TeB~\cite{teb} & {224 $\times$ 224 }    & {Resnet 101} & 57.0 &67.2 &56.1 &54.3 &58.7 &57.3 &68.5 &61.5 &56.3 &60.9 \\
 
SAGNN~\cite{sagnet}  & {473 $\times$ 473 }    & {Resnet 50} & 64.7 &69.6 &57.0 &57.2 &62.1   & 64.9  &70.0  &57.0  &59.3  &62.8  \\ 

ASGNet~\cite{asgnet} & {473 $\times$ 473 }    & {Resnet 50} &58.8 &67.8 &56.8 &53.6 &59.3 &63.6 &70.5 &64.1 &57.4 &63.9 \\

ASR~\cite{ASR} & {473 $\times$ 473 }    & {Resnet 50} &55.2 &70.3 &53.4 &53.6 &58.1 &59.4 &71.8 &56.8 &55.7 &61.0   \\

RePRI~\cite{citeb} & {417 $\times$ 417 }    & {Resnet 50} & 60.2  &67.0  &61.7  &47.5  &59.1  &64.5  &70.8  &71.7  &60.3  &66.8 \\ 

HSNet~\cite{citec} & {473 $\times$ 473 }    & {Resnet 50} & 64.3  &70.7  &60.3  &60.5  &64.0  & 70.3  &73.2  &67.4  &67.1  &69.5 \\ 

MLC~\cite{cited} & {473 $\times$ 473 }    & {Resnet 50} & 59.2 &71.2 &65.6 &52.5 &62.1 &63.5 &71.6 &71.2 &58.1 &66.1 \\ 

PFENet~\cite{pfenet} & {473 $\times$ 473 }    & {Resnet 50} & 61.7   & 69.5   & 55.4   & 56.3   & 60.8 & 63.1   & 70.7   & 55.8   & 57.9   & 61.9 \\ 

\hline
CRNet$^*$~\cite{crnet} & {321 $\times$ 321 }    & {Resnet 50} & 56.8   & 65.8   & 49.4   & 50.6   & 55.7 & 58.7   & 67.9   & 54.2   & 53.5   & 58.8 \\

CRCNet  & {321 $\times$ 321 }    & {Resnet 50} & {57.1}   & 65.7   & {51.4}   & 51.2   & {56.4} & {58.9}   & {68.1}   & 55.1   & {54.1}   & {59.1} \\

PFENet + GC & {473 $\times$ 473 }    & {Resnet 50} &61.9	&69.5	&55.2	&56.9	&60.9  &62.8	&70.9	&55.9	&61.5	&62.8 \\
PFENet + CR & {473 $\times$ 473 }    & {Resnet 50} &63.5	&69.6	&55.4	&57.0	    &61.4  &64.8	&71.0	    &57.5	&60.6	&63.4 \\
PFENet + CRC & {473 $\times$ 473 }    & {Resnet 50}  &63.4	&69.7	&55.8	&56.9   &61.5 &65.2	&70.9	&55.9	&61.8	&63.5 \\

\bottomrule[1pt]
\end{tabular}%
}

\label{Table:voc-soa-miou}
\end{table*}

\begin{table}[]
\centering
\small
\caption{Comparison with the state-of-the-art methods under the 1-shot and 5-shot settings. Our proposed network outperforms all previous methods and achieves new state-of-the-art performance. The results are reported on PASCAL VOC 2012 dataset with FBIoU(\%). $^*$ denotes the reproduced results. PFENet denotes that we build our method upon the PFENet~\cite{pfenet}. The CRC denotes utilizing both our conditional module and the cross-reference module.}
\resizebox{0.7\columnwidth}{!}
{
\begin{tabular}{lcc}
\toprule[1pt]
Method & 1-shot (\%) & 5-shot (\%) \\
\hline
OSLM \cite{shaban2017one}       & 61.3    & 61.5          \\
co-fcn \cite{rakelly2018conditional}   & 60.9      & 60.2          \\
sg-one  \cite{zhang2018sg}      & 63.1  & 65.9          \\
R-DFCN \cite{siam2019adaptive}   & 60.9   & 66.0          \\
PL   \cite{Dong2018FewShotSS}         & 61.2     & 62.3          \\
A-MCG    \cite{hu2019attention}      & 61.2    & 62.2          \\
CANet \cite{zhang2019canet}        & 66.2    & 69.6          \\
SAGNN~\cite{sagnet} & 73.2    & 73.3         \\

ASGNet~\cite{asgnet} & 69.2    & 74.2          \\

ASR~\cite{ASR} & 71.3    & 72.5          \\

PFENet~\cite{pfenet} & 71.3    & 72.5          \\
\hline
CRNet$^*$~\cite{crnet}      & 66.8    & 71.5     \\

CRCNet     & {66.9}    & {71.7}     \\

FPENet + CRC  & {71.4}    & {73.4} \\
\bottomrule[1pt]
\end{tabular}
}
\label{Table: voc-soa-fbiou}
\end{table}
\section{Experiment}

\begin{table}[]
\centering
\small
\caption{The class division of the  PASCAL VOC 2012 dataset proposed in~\cite{shaban2017one}. }
\resizebox{0.95\columnwidth}{!}
{
\begin{tabular}{c|c}
\toprule[1pt]
fold & categories \\\hline
0 & aeroplane, bicycle, bird, boat, bottle  \\ \hline
1 & bus, car, cat, chair, cow        \\\hline
2 & diningtable, dog, horse, motobike, person          \\\hline
3 &  potted plant, sheep, sofa, train, tv/monitor       \\
\bottomrule[1pt]
\end{tabular}%
}
\label{division}
\end{table}
\subsection{Implementation Details}
In the Siamese encoder, we exploit multi-level features from the ImageNet pre-trained Resnet-50 as the image representations. We use dilated convolutions and keep the feature maps after layer3 and layer4 have a fixed size of 1/8 of the input image and concatenate them for final prediction. 
All the convolutional layers in our proposed modules have the kernel size of $3\times3$ and generate features of 256 channels, followed by the ReLU activation function. 
Our network is end-to-end training. 
We utilize SGD as the optimizer to train the network for 200 epochs with a Binary cross-entropy loss. The learning rate is set to 0.0025.  We apply the random crop, random scale, and random flip for data augmentation during the training process. 
We adopt a multi-scale strategy with scales [0.75,1.25] and fuse all predictions during the inference phase. 
We recurrently run the mask refinement module 10 times to refine the predicted masks. In the case of $k$-shot learning, we fix the Siamese encoder and finetune the rest parameters.

\subsection{Evaluation Metric and Datasets}
\textbf{Evaluation Metric.}
In previous works, there exist two evaluation metrics. Shaban\etal~\cite{shaban2017one} report the results with the mean of per-class foreground Intersection-Over-Union(mIoU). While~\cite{Dong2018FewShotSS,rakelly2018conditional} ignore the categories and report the results with the mean of foreground IoU and background IoU (FBIoU). We choose the standard mIoU as our evaluation metric for the following two reasons: 1) The unbalanced image distribution (class sheep contains 49 images while the class person contains 378 images). 2) The background IoU is very high for small objects, which will fail to evaluate the model's capability. Nevertheless, we still compare the previous state-of-the-art methods with both evaluation metrics. The evaluation metrics are calculated as follows:

\begin{equation}
    IoU = \frac{Intersection}{Union} = \frac{TP}{TP + FP + FN}, 
    \label{equal:iou}
\end{equation}

\begin{equation}
    mIoU = \frac{1}{n} \sum_{1}^{n}(IoU_{n}),
\end{equation}

\begin{equation}
    FBIoU = \frac{1}{2}(IoU_{fg} + IoU_{bg}).
\end{equation}
The TP denotes true positive, FP denotes false positive, FN denotes false negative, $n$ denotes the classes' number. The standard mIoU is calculated by averaging the IoU of all classes. The $IoU_{fg}$ is calculated with equation~\ref{equal:iou}, which ignores the object categories, and $IoU_{bg}$ is calculated in the same way but reversed the foreground and background. FBIoU average the $IoU_{fg}$ and the $IoU_{bg}$.

\textbf{PASCAL VOC 2012.}
We implement cross-validation experiments on the PASCAL VOC 2012 dataset. To compare our model with previous works, we adopt the same category divisions and test settings, which are first proposed in~\cite{shaban2017one}. In the cross-validation experiments, 20 object categories are evenly divided into 4 folds, with three folds as the training classes and one fold as the testing classes. The category division is shown in Table~\ref{division}. We report the average performance over 4 testing folds. We use the standard mean Intersection-over-Union (mIoU) of the classes in the testing fold for the evaluation metrics. For more details about the dataset information and the evaluation metric, please refer to~\cite{shaban2017one}. The PASCAL VOC dataset is publicly available at \url{http://host.robots.ox.ac.uk/pascal/VOC}. 

\textbf{MS COCO and FSS-1000.}
One limitation of PASCAL VOC 2012 is that it involves only a few categories, which is insufficient to verify the model's capabilities on the few shot segmentation tasks.
To alleviate this problem, we implement experiments on larger datasets that contain more categories and images, such as MS COCO and FSS-1000 to validate our network. The MS COCO dataset is publicly available at \url{https://cocodataset.org/#home} and the FSS-1000 dataset is publicly available at \url{https://github.com/HKUSTCV/FSS-1000}.

COCO 2014~\cite{coco} is a challenging large-scale dataset containing 80 object categories and 82783 training images, and 40504 validation images. 
To validate our method, we follow two few shot MC COCO dataset settings. 
Following~\cite{zhang2019canet}, we choose 40 classes for training, 20 classes for validation, and 20 classes for the test. We report the average standard mean Intersection-over-Union (mIoU) of the classes in the testing set for the evaluation metrics. 

We also conduct the experiments on few shot MS COCO same setting with ~\cite{sagnet,asgnet,ASR,pfenet}, as shown in Table~\ref{Table:coco-soa-new} and Table~\ref{table:coco-bfiou}, we report the results with both standard mIoU and FBIoU metrics.

FSS-1000 introduces a more realistic dataset for the few shot segmentation tasks by increasing the number of object categories instead of the number of images. Particularly, FSS-1000 contains 1000 classes, and every class includes 10 images. Following~\cite{fss1000}, we choose 520 classes for training, 240 classes for validation and 240 classes for testing.
We adopt the mean Intersection-over-Union (mIoU) of the classes as the evaluation metrics. 

\subsection{Ablation study} \label{ablationsec}
The goal of the ablation study is to inspect each component in our network design. 
Our ablation experiments are conducted on the PASCAL VOC 2012 dataset.
We implement cross-validation 1-shot and k-shot experiments and report the average performance over the four splits on the PASCAL VOC 2012.
In Table~\ref{Table:ablation-condition-cr}, we first investigate the contributions of our two essential network components: the conditional module and the cross-reference module. As shown, there are significant performance drops if we remove either component from the network. Our network can improve the counterpart model without the cross-reference module. To further investigate the contributions of the network components to other methods, we entangled the component of our CRCNet into PFENet~\cite{pfenet}, and conducted the experiments on both PASCAL VOC 2012 and MS COCO datasets. The results are shown in Table~\ref{Table:voc-soa-miou} and Table~\ref{Table:coco-soa-new}. 

We adopt a multi-scale test experiment to investigate how much the objects' scale variance influences network performance. Specifically, we resize the support image and the query image to [0.75,1,1.25] of the original image size and conduct the test time inference. The output predicted mask of the resized query image is bilinearly resized to the original image size. We fuse the predictions under different image scales. As shown in Table~\ref{table:abalation-refine-scale}, multi-scale input test brings 0.8 mIoU score  improvement in the 1-shot setting. 
We also investigate the choices of features in the network backbone in Table~\ref{table:abalation-refine-scale}. We compare the multi-level feature embedding with the features solely from the last layer. Our model with multi-level features provides an improvement of 3.4 mIoU score, which indicates that the middle-level features are also relevant and helpful in locating the common objects in two images better.

\begin{table}[]
\centering
\small
\caption{Comparison with the state-of-the-art methods under the 1-shot and 5-shot settings. Our proposed network outperforms all previous methods and achieves new state-of-the-art performance. The results are reported on the MS COCO dataset with FBIoU(\%). $^*$ denotes the reproduce results. PFENet denotes that we build our method upon the PFENet~\cite{pfenet}. The CRC denotes utilizing both our conditional module and the cross-reference module.}
\resizebox{0.7\columnwidth}{!}
{
\begin{tabular}{lcc}
\toprule[1pt]
Method & 1-shot (\%) & 5-shot (\%) \\
\hline
DAN \cite{danet}   & 62.3     & 60.2          \\

SAGNN~\cite{sagnet} & 60.4    & 67.0         \\

ASGNet~\cite{asgnet} & 60.0    & 63.4        \\

PFENet~\cite{pfenet} &58.6   & 61.9  \\

\hline
PFENet + CRC     & {62.0}    & {66.1}     \\

\bottomrule[1pt]
\end{tabular}
}
\label{table:coco-bfiou}
\end{table}

\begin{table*}[]
\small
\centering
\caption{Detailed 1-shot and 5-shot results on each split under the mIoU(\%) evaluation metric with MS COCO dataset. Our method outperforms most state-of-the-art methods and achieves comparable results with the most recent work. PFENet denotes that we build our method upon the PFENet~\cite{pfenet}. The GC denotes the global conditional module. The CR denotes the cross-reference module, and the CRC denotes utilizing both our conditional module and the cross-reference module.
}

\resizebox{1\textwidth}{!}{%
\begin{tabular}{l|c|c|ccccc|ccccc}
\toprule[1pt]

\multirow{2}{*}{Method} & \multirow{2}{*}{Training Size} & \multirow{2}{*}{Backbone} & \multicolumn{5}{c|}{1-shot} & \multicolumn{5}{c}{5-shot} \\ \cline{4-13} 
 & & & split-0 & split-1 & split-2 & split-3 & mean & split-0 & split-1 & split-2 & split-3 & mean \\ \hline\hline

PANet~\cite{wang2019panet}  & {417 $\times$ 417 }    & {VGG 16} & -- & -- & -- & -- & 20.9 & -- & -- & -- & -- & 29.7   \\
FWB\cite{fast-weight} & {512  $\times$ 512}                       & {VGG 16}  & 16.9 & 17.9 & 20.9 & 28.8 & 21.2 & 19.1 & 21.4 & 23.9 & 30.0 & 23.6 \\

RPMMs\cite{pmm}     & {321 $\times$ 321 }    & {Resnet 50}   & 29.5   & {36.8}   & 28.9   & 27.0   & 30.5  & 33.8   & {41.9}   & 32.9   & 33.3   & 35.5 \\

CANet* \cite{zhang2019canet} & {321 $\times$ 321 }    & {Resnet 50}  & 25.1  & 30.3                           & 24.5 & 24.7  & 26.1    & 26.0 & 32.4 & 26.1 & 27.0 & 27.9 \\ 

PPNet~\cite{ppnet}    & {417 $\times$ 417 }    & {Resnet 50}    & 28.1 & 30.8 & 29.5 & 27.7 & 29.0 & 39.0 & 40.8 & 37.1 & 37.3 & 38.5  \\

RePRI~\cite{citeb} & {417 $\times$ 417 }    & {Resnet 50} & 31.2 &38.1 &33.3 &33.0 &34.0 &38.5 &46.2 &40.0 &43.6 &42.1 \\ 

HSNet~\cite{citec} & {473 $\times$ 473 }    & {Resnet 50} &  36.3 &43.1 &38.7 &38.7 &39.2 &43.3 &51.3 &48.2 &45.0 &46.9 \\ 

MLC~\cite{cited} & {473 $\times$ 473 }    & {Resnet 50} & 46.8  &35.3  &26.2  &27.1  &33.9 & 54.1  &41.2  &34.1  &33.1  &40.6 \\ 

SAGNN~\cite{sagnet}  & {473 $\times$ 473 }    & {Resnet 50}   &36.1 &41.0 &38.2 &33.5 &37.2  &40.9 &48.3 &42.6 &38.9 &42.7 \\ 

ASGNet~\cite{asgnet} & {473 $\times$ 473 }    & {Resnet 50}  &- &-  &- &-  &34.5  &- &- &- &- &42.5 \\

ASR~\cite{ASR} & {473 $\times$ 473 }    & {Resnet 50}   &29.9 &35.0 &31.8 &33.5 &32.5 &31.2 &37.8 &33.4 &35.2 &34.3  \\

PFENet~\cite{pfenet} & {473 $\times$ 473 }    & {Resnet 50}  & 34.3   & 33.0   & {32.3}   & 30.1   & 32.4  & 38.5   & 38.6   & {38.2}   & {34.3}   & 37.4 \\
\hline

PFENet + CR & {473 $\times$ 473 }    & {Resnet 50}   &34.9	&41.4	&38.8	&36.2	&37.8		&37.9	&46.3	&42.6	&39.6	&41.6  \\ 

PFENet + GC   & {473 $\times$ 473 }    & {Resnet 50}  &34.8	&41.8	&40.2	&36.0	&38.2		&38.8	&45.3	&42.2	&41.0	&41.8   \\ 

PFENet + CRC   & {473 $\times$ 473 }    & {Resnet 50}  &35.1	&42.2	&41.3	&36.4	&38.7		&41.1	&46.5	&43.8	&41.3	&43.1   \\ 

\bottomrule[1pt]
\end{tabular}%
}
\label{Table:coco-soa-new}
\end{table*}

\begin{table}[]
\small
\centering
\caption{
Comparison with the state-of-the-art methods on MS COCO dataset. Our model outperforms all previous methods and achieves new state-of-the-art performance under both 1-shot setting and 5-shot setting. MR denotes Mask Refine module, MR-CR denotes Mask Refine module and Cross-Reference module, MS denotes Multi-Scale. All our modules bring a performance improvement over the baseline model(CRCNet w/o MR-CR). The results are reported with standard mIoU(\%).} 

\resizebox{0.9\columnwidth}{!}{%
\begin{tabular}{lcc|cc}
\toprule[1pt]

\multirow{2}{*}{Method} & \multicolumn{2}{c|}{1-shot} & \multicolumn{2}{c}{5-shot} \\ \cline{2-5} 
 & MS & mIoU & MS & mIoU \\ \hline
 
CANet~\cite{zhang2019canet} &         &46.3            &   &49.7               \\ 
CANet~\cite{zhang2019canet} &  \checkmark & 49.9      &  \checkmark & 51.6            \\ \hline

CRCNet w/o MR-CR &       &   45.4    &     & 46.1  \\
CRCNet  w/o MR &       & 47.2    &       & 47.9  \\
CRCNet  &       & 49.4    &       & 50.6  \\
CRCNet  & \checkmark & \textbf{51.2}   & \checkmark & \textbf{52.9} \\

\bottomrule[1pt]
\end{tabular}%
}
\label{coco-soa}
\end{table}

 \begin{figure*}[t]
  \centering
    \includegraphics[width=1\linewidth]{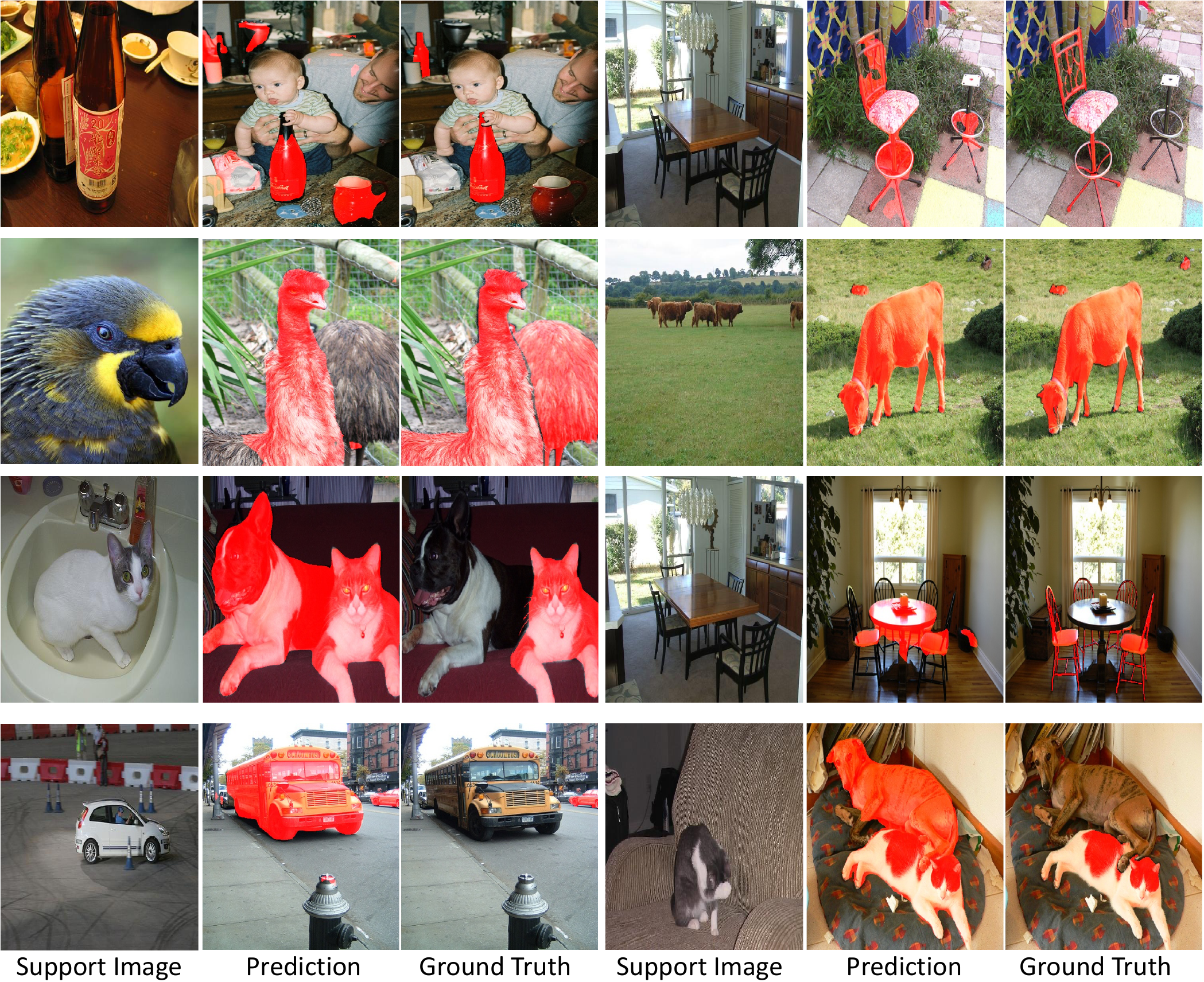}
    \caption{The failure cases on PASCAL VOC 2012 dataset.}
    \label{Figure:fail_voc}
\end{figure*}

 \begin{figure*}[t]
  \centering
    \includegraphics[width=1\linewidth]{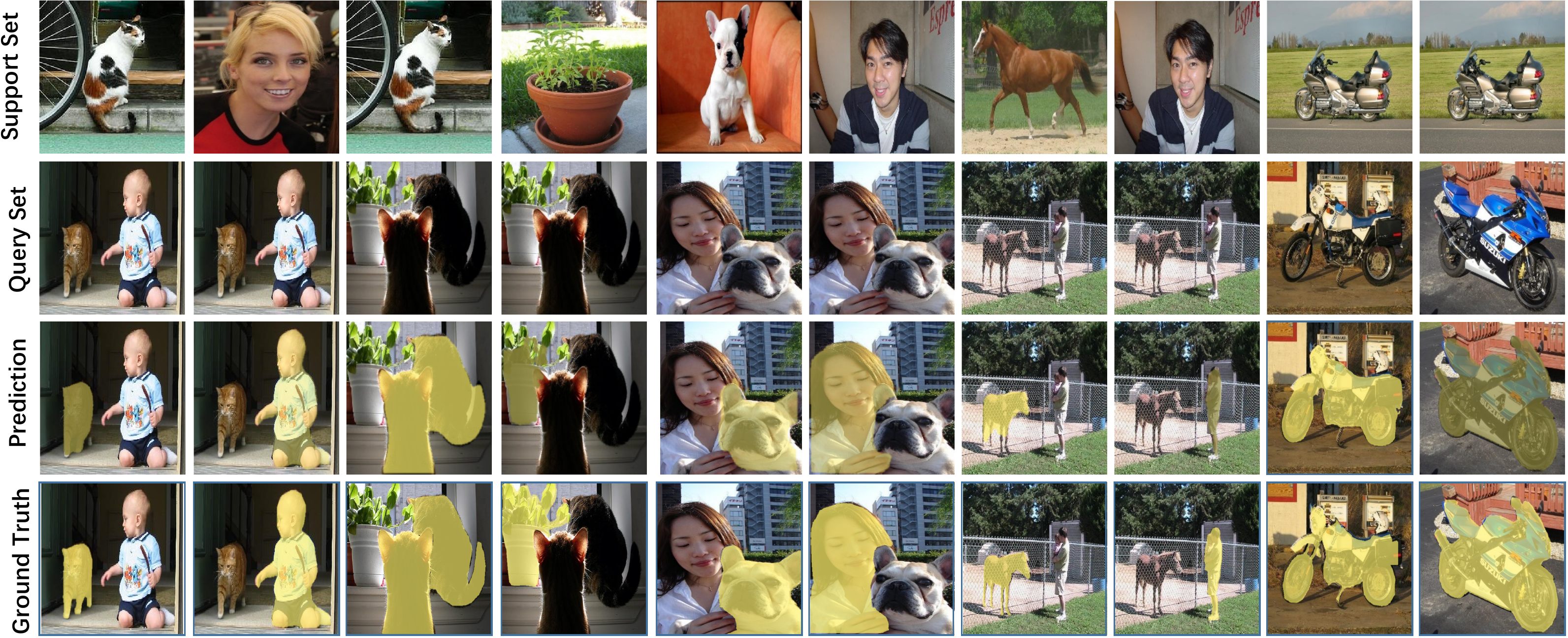}
    \caption{Our Qualitative examples on the PASCAL VOC dataset. The first row is the support set, and the second row is the query set. The third row is our predicted results, and the fourth row is the ground truth. Even when the query images contain objects from multiple classes, our network can still successfully segment the target category indicated by the support mask. }
    \label{Figure:results-voc}
\end{figure*}

 \begin{figure*}[t]
  \centering
    \includegraphics[width=1\linewidth]{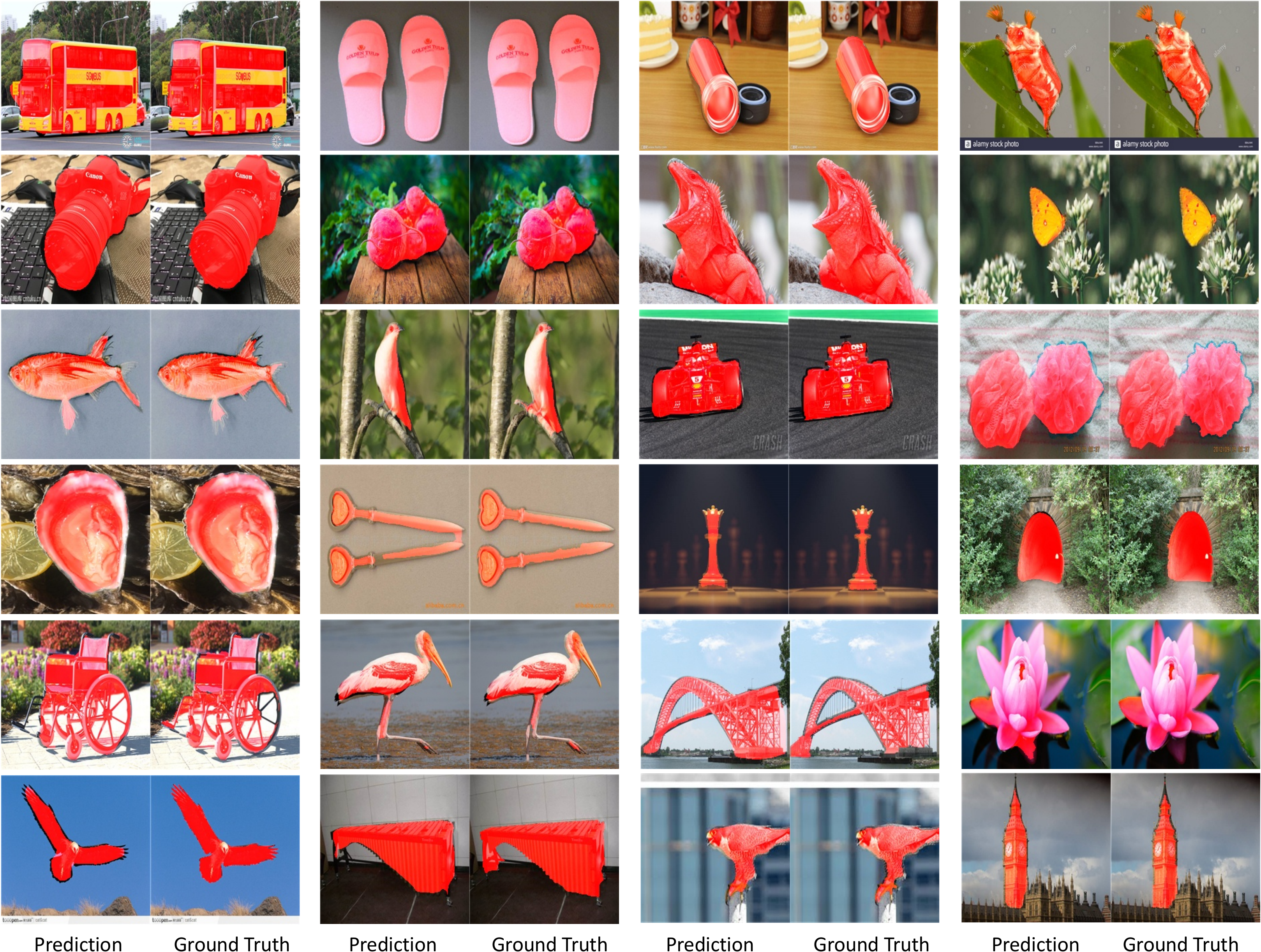}
    \caption{Our qualitative examples on the FSS-1000 dataset. The left column is the prediction, and the right column is the ground truth.}
    \label{Figure:results-fss}
\end{figure*}

 \begin{figure*}[t]
  \centering
    \includegraphics[width=1\linewidth]{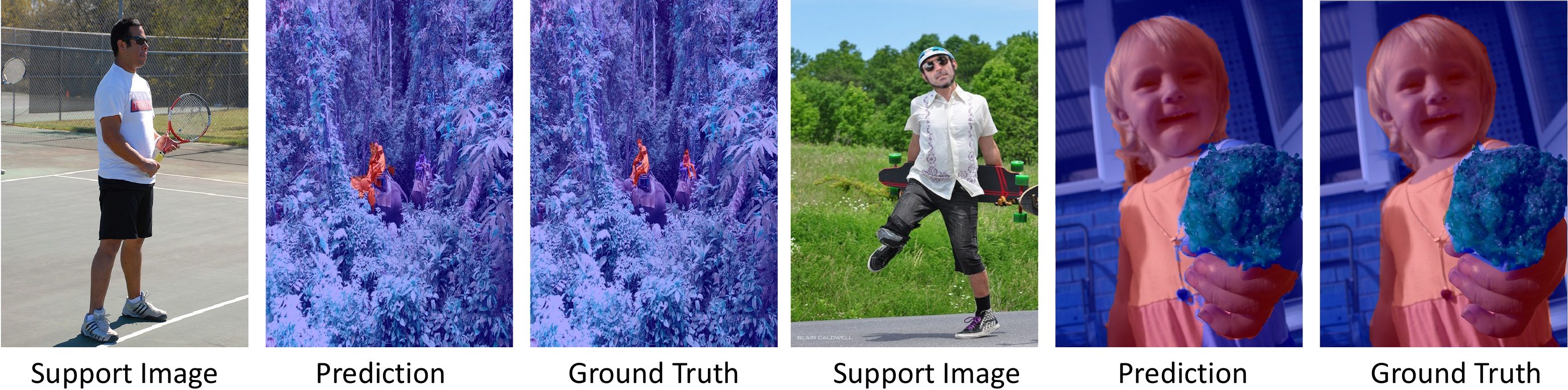}
    \caption{Our qualitative examples on the COCO dataset.}
    \label{Figure:results-coco}
\end{figure*}

To further inspect the mask refinement module's effectiveness, we design a baseline model that removes the cached branch. In this case, the mask refinement block makes predictions solely based on the input features, and we only run the mask refinement module once.
As shown in Table~\ref{table:abalation-refine-scale} , our mask refinement module brings
2.6 mIoU score performance increased over our baseline method.

To further investigate how the global conditional and local conditional modules influence the network performance, we implement cross-validation 1-shot experiments on PASCAL VOC 2012. As shown in Table~\ref{ablition-condition}, there is a significant performance increase if we only add the global conditional or the local conditional module to the baseline. Moreover, if we combine both global and local representations as our condition, the combined state brings 4.4 mIoU improvement.

In the $k$-shot setting, we compare our finetuning-based method with the fusion-based methods widely used in previous activities. We make an inference with each of the support images and average their probability maps as final predictions for the fusion-based method. The comparison is shown in Table~\ref{ablation:Fuse-and-FT}. In the 5-shot setting, the finetuning-based method outperforms the 1-shot baseline by 8.4 mIoU score, which is significantly superior to the fusion-based method. When 10 support images are available, our finetuning-based method shows more advantages. The performance continues increasing while the fusion-based method's performance begins to drop. As shown in Table~\ref{Table:PFE-FT}, we also adapt the finetuning scheme to PFENet~\cite{pfenet}, which brings the improvements to the 5-shot setting on PASCAL VOC 2012 dataset.

To investigate the effectiveness of different activation functions in the cross-reference module such as Relu, Sigmoid, and Softmax. We have conducted the experiment with different activation functions. The results are shown in Table~\ref{Table: abalation-sigmoid}. 
We also conducted experiments on how the effectiveness of the support mask in the local conditional module, which aims to suppress the background information for the support features. The results are shown in Table~\ref{Table: abalation-local-condition-mask}.

\subsection{Weakly supervised few shot segmentation}
Following CANet~\cite{zhang2019canet} and TeB~\cite{teb}, we further evaluate our CRCNet with a weaker annotation such as bounding boxes.
We generate the bounding box annotations from PASCAL VOC 2012, SDS~\cite{hariharan2014simultaneous}, and FSS-1000~\cite{fss1000} dataset by locating every object. 
As is shown in Table~\ref{weakly-1-shot}, our CRCNet achieves comparable performances with bounding box annotation to the results with pixel-wise supervision under both PASCAL VOC 2012 and FSS-1000 dataset, which indicate our CRCNet is robust to the annotation with noisy background areas within the bounding boxes. 
Furthermore, our CRCNet outperforms all the previous methods in both PASCAL VOC 2012 and FSS-1000 datasets under the weakly supervised few shot segmentation setting.

\begin{table}[]
\centering
\small
\caption{Comparison with the state-of-the-art methods under the weakly supervised few-shot setting. Our proposed network achieves state-of-the-art performance. The results are reported on PASCAL VOC 2012 dataset and FSS-1000 dataset with standard mIoU(\%). }
\resizebox{1\columnwidth}{!}
{
\begin{tabular}{l|c|c}
\toprule[1pt]
Method                           & \multicolumn{2}{c}{mIoU(\%)} \\ \hline
                                 & PASCAL     & FSS-1000    \\ \hline
CANet~\cite{zhang2019canet} (Bounding box annotations) & 52.0       &   -         \\
TeB~\cite{teb} (Bounding box annotations) & -       &   78.2         \\ \hline
CRCNet (Bounding box annotations) & \textbf{53.4}       & \textbf{79.9}        \\ 
\bottomrule[1pt]
\end{tabular}
}
\label{weakly-1-shot}
\end{table}

\subsection{Comparison with the State-of-the-Art Results}
We compare our network with state-of-the-art methods on the PASCAL VOC 2012 dataset, MS COCO dataset, and FSS-1000 dataset.

\textbf{PASCAL VOC 2012.}
Table~\ref{Table:voc-soa-miou} and Table~\ref{Table: voc-soa-fbiou} shows the performance of different methods in both 1-shot and 5-shot setting. We use FBIoU to denote the evaluation metric proposed in~\cite{rakelly2018conditional} and mIoU to denote the standard mean Intersection-Over-Union. The difference between the two metrics is that the FBIoU metric also incorporates the background into the Intersection-over-Union computation and ignores the image category. We achieve the new state-of-the-art under both the 1-shot and 5-shot settings with both evaluation metrics. \\

\textbf{FSS-1000.} The comparison results with FSS-1000 dataset~\cite{fss1000} under 1-shot and 5-shot setting is shown in Table~\ref{fss-results}. Our CRCNet outperforms the previous state-of-the-art methods by 5.6 (resp. 2.0) mIoU under 1-shot (resp. 5-shot) setting, and achieves new state-of-the-art performance. 
\begin{table}[]
\centering
\small
\caption{Comparison with the state-of-the-art methods under the 1-shot and 5-shot settings with FSS-1000 dataset. Our proposed network achieves state-of-the-art performance. The results are reported with standard mIoU(\%). $^*$ denotes the reproduce results.
}
\resizebox{0.6\columnwidth}{!}
{
\begin{tabular}{l|l|l}
\toprule[1pt]
Method       & 1-shot        & 5-shot        \\
\hline
OSLSM~\cite{shaban2017one}          & 70.3          & 73.0          \\
co-fcn~\cite{rakelly2018conditional}        & 71.2          & 74.2          \\
FSS-1000~\cite{fss1000}      & 73.4          & 80.1          \\
FOMAML~\cite{fomaml}       & 75.1          & 80.6          \\
\hline
CRNet$^*$~\cite{crnet} & 80.5 & 82.3  \\
CRCNet & \textbf{81.1} & \textbf{83.2}  \\

\bottomrule[1pt]
\end{tabular}
}
\label{fss-results}
\end{table}

\textbf{MS COCO.}
As mentioned above, to validate our method, we conduct the experiments following two few-shot MC COCO dataset settings. 
The comparison results with few-shot MS COCO dataset same setting with CANet~\cite{zhang2019canet} under 1-shot and 5-shot settings are shown in Table~\ref{coco-soa}. Our CRCNet outperforms all previous state-of-the-art methods and achieves new state-of-the-art performance. 
In particular, the ablation study on MS COCO still draws the same conclusion as the PASCAL VOC 2012. In particular, our CRCNet achieves 5.8 mIoU improvement over the baseline (conditional module only) under the 1-shot setting and 6.8 mIoU improvement under the 5-shot setting.

We also the experiments on MS COCO same setting with ~\cite{sagnet,asgnet,ASR}, as shown in Table~\ref{Table:coco-soa-new} and Table~\ref{table:coco-bfiou}, we report the result with both standard mIoU and FBIoU metrics.

\subsection{Failure case analysis}
In this section, we analyze the challenging cases that fail our model. As shown in Figure~\ref{Figure:fail_voc}, our model fails to distinguish between the dogs and cats, chairs and tables, cars and buses, teacups and bottles. This is because they have a similar pattern: difficult to distinguish without semantic information. Our model fails to find the small cow possible because the small object is difficult to capture without the low-level features. Moreover, our model can not locate the bird bodies because the support image only contains the bird's head information. 
\section{Conclusion}
In this paper, we have presented a novel cross-reference with a local-global conditional network for few-shot segmentation. Unlike previous work unilaterally guiding the segmentation of query images with support images, our two-head design concurrently makes predictions in both the query image and the support image to help the network better locate the target category. We develop a mask refinement module with a cache mechanism that can effectively improve the prediction performance. In the $k$-shot setting, our finetuning-based method can significantly improve performance by taking advantage of more annotated data. Extensive ablation experiments on PASCAL VOC 2012, FSS-1000, and MS COCO datasets validate the effectiveness of our design. Our model achieves state-of-the-art performance on the PASCAL VOC 2012 dataset, MS COCO dataset, and FSS-1000 dataset. 
\section*{Acknowledgements} This research is supported by the National Research Foundation, Singapore under its AI Singapore Programme (AISG Award No: AISG-RP-2018-003), the Ministry of Education, Singapore, under its Academic Research Fund Tier 2 (MOE-T2EP20220-0007) and Tier 1 (RG95/20).
This research is also partly supported by the Agency for Science, Technology and Research (A*STAR) under its AME Programmatic Funds (Grant No. A20H6b0151).

\bibliographystyle{spmpsci}
\bibliography{egbib}

\begin{thebibliography}{10}
\providecommand{\url}[1]{{#1}}
\providecommand{\urlprefix}{URL }
\expandafter\ifx\csname urlstyle\endcsname\relax
  \providecommand{\doi}[1]{DOI~\discretionary{}{}{}#1}\else
  \providecommand{\doi}{DOI~\discretionary{}{}{}\begingroup
  \urlstyle{rm}\Url}\fi

\bibitem{teb}
Azad, R., Fayjie, A.R., Kauffmann, C., Ben~Ayed, I., Pedersoli, M., Dolz, J.:
  On the texture bias for few-shot cnn segmentation.
\newblock In: Proceedings of the IEEE/CVF Winter Conference on Applications of
  Computer Vision, pp. 2674--2683 (2021)

\bibitem{citeb}
Boudiaf, M., Kervadec, H., Masud, Z.I., Piantanida, P., Ben~Ayed, I., Dolz, J.:
  Few-shot segmentation without meta-learning: A good transductive inference is
  all you need?
\newblock In: Proceedings of the IEEE/CVF Conference on Computer Vision and
  Pattern Recognition, pp. 13979--13988 (2021)

\bibitem{chen2018semantic}
Chen, H., Huang, Y., Nakayama, H.: Semantic aware attention based deep object
  co-segmentation.
\newblock arXiv preprint arXiv:1810.06859  (2018)

\bibitem{chen2018deeplab}
Chen, L.C., Papandreou, G., Kokkinos, I., Murphy, K., Yuille, A.L.: Deeplab:
  Semantic image segmentation with deep convolutional nets, atrous convolution,
  and fully connected crfs.
\newblock IEEE transactions on pattern analysis and machine intelligence
  \textbf{40}(4), 834--848 (2018)

\bibitem{review3_1}
Chen, W., Jiang, Z., Wang, Z., Cui, K., Qian, X.: Collaborative global-local
  networks for memory-efficient segmentation of ultra-high resolution images.
\newblock In: Proceedings of the IEEE/CVF Conference on Computer Vision and
  Pattern Recognition, pp. 8924--8933 (2019)

\bibitem{chen2019closerfewshot}
Chen, W.Y., Liu, Y.C., Kira, Z., Wang, Y.C., Huang, J.B.: A closer look at
  few-shot classification.
\newblock In: International Conference on Learning Representations (2019)

\bibitem{review3_3}
Chen, Y., Cao, Y., Hu, H., Wang, L.: Memory enhanced global-local aggregation
  for video object detection.
\newblock In: Proceedings of the IEEE/CVF Conference on Computer Vision and
  Pattern Recognition, pp. 10337--10346 (2020)

\bibitem{imagenet}
Deng, J., Dong, W., Socher, R., Li, L.J., Li, K., Fei-Fei, L.: Imagenet: A
  large-scale hierarchical image database.
\newblock In: Proceedings of the IEEE/CVF Conference on Computer Vision and
  Pattern Recognition, pp. 248--255 (2009)

\bibitem{Dong2018FewShotSS}
Dong, N., Xing, E.: Few-shot semantic segmentation with prototype learning.
\newblock In: BMVC, p.~79 (2018)

\bibitem{mdl}
Dong, Z., Zhang, R., Shao, X., Zhou, H.: Multi-scale discriminative
  location-aware network for few-shot semantic segmentation.
\newblock In: 2019 IEEE 43rd Annual Computer Software and Applications
  Conference (COMPSAC), vol.~2, pp. 42--47. IEEE (2019)

\bibitem{review3_4}
Fan, Q., Zhuo, W., Tang, C.K., Tai, Y.W.: Few-shot object detection with
  attention-rpn and multi-relation detector.
\newblock In: Proceedings of the IEEE/CVF Conference on Computer Vision and
  Pattern Recognition, pp. 4013--4022 (2020)

\bibitem{tip-co-seg1}
Han, J., Quan, R., Zhang, D., Nie, F.: Robust object co-segmentation using
  background prior.
\newblock IEEE Transactions on Image Processing \textbf{27}(4), 1639--1651
  (2017)

\bibitem{hariharan2014simultaneous}
Hariharan, B., Arbel{\'a}ez, P., Girshick, R., Malik, J.: Simultaneous
  detection and segmentation.
\newblock In: European Conference on Computer Vision, pp. 297--312. Springer
  (2014)

\bibitem{he2016deep}
He, K., Zhang, X., Ren, S., Sun, J.: Deep residual learning for image
  recognition.
\newblock In: Proceedings of the IEEE conference on computer vision and pattern
  recognition, pp. 770--778 (2016)

\bibitem{fomaml}
Hendryx, S.M., Leach, A.B., Hein, P.D., Morrison, C.T.: Meta-learning
  initializations for image segmentation.
\newblock arXiv preprint arXiv:1912.06290  (2019)

\bibitem{hou2022distilling}
Hou, J., Ding, H., Lin, W., Liu, W., Fang, Y.: Distilling knowledge from object
  classification to aesthetics assessment.
\newblock IEEE Transactions on Circuits and Systems for Video Technology
  (2022)

\bibitem{hou2022interaction}
Hou, J., Lin, W., Yue, G., Liu, W., Zhao, B.: Interaction-matrix based
  personalized image aesthetics assessment.
\newblock IEEE Transactions on Multimedia  (2022)

\bibitem{review3_2}
Hou, R., Chang, H., Ma, B., Shan, S., Chen, X.: Cross attention network for
  few-shot classification.
\newblock arXiv preprint arXiv:1910.07677  (2019)

\bibitem{hu2019attention}
Hu, T., Yang, P., Zhang, C., Yu, G., Mu, Y., Snoek, C.G.: Attention-based
  multi-context guiding for few-shot semantic segmentation.
\newblock In: Proceedings of the AAAI conference on artificial intelligence,
  pp. 8441--8448 (2019)

\bibitem{joulin2012multi}
Joulin, A., Bach, F., Ponce, J.: Multi-class cosegmentation.
\newblock In: 2012 IEEE Conference on Computer Vision and Pattern Recognition,
  pp. 542--549. IEEE (2012)

\bibitem{kolesnikov2016seed}
Kolesnikov, A., Lampert, C.H.: Seed, expand and constrain: Three principles for
  weakly-supervised image segmentation.
\newblock In: European Conference on Computer Vision, pp. 695--711. Springer
  (2016)

\bibitem{krizhevsky2012imagenet}
Krizhevsky, A., Sutskever, I., Hinton, G.E.: Imagenet classification with deep
  convolutional neural networks.
\newblock In: Advances in neural information processing systems, pp. 1097--1105
  (2012)

\bibitem{asgnet}
Li, G., Jampani, V., Sevilla{-}Lara, L., Sun, D., Kim, J., Kim, J.: Adaptive
  prototype learning and allocation for few-shot segmentation.
\newblock In: Proceedings of the IEEE/CVF Conference on Computer Vision and
  Pattern Recognition, pp. 8334--8343 (2021)

\bibitem{fss1000}
Li, X., Wei, T., Chen, Y.P., Tai, Y.W., Tang, C.K.: Fss-1000: A 1000-class
  dataset for few-shot segmentation.
\newblock In: Proceedings of the IEEE/CVF Conference on Computer Vision and
  Pattern Recognition, pp. 2869--2878 (2020)

\bibitem{refine-tpami}
Lin, G., Liu, F., Milan, A., Shen, C., Reid, I.: Refinenet: Multi-path
  refinement networks for dense prediction.
\newblock IEEE transactions on pattern analysis and machine intelligence
  \textbf{42}(5), 1228--1242 (2019)

\bibitem{lin2017refinenet}
Lin, G., Milan, A., Shen, C., Reid, I.D.: Refinenet: Multi-path refinement
  networks for high-resolution semantic segmentation.
\newblock In: Proceedings of the IEEE/CVF Conference on Computer Vision and
  Pattern Recognition, p.~5 (2017)

\bibitem{coco}
Lin, T.Y., Maire, M., Belongie, S., Hays, J., Perona, P., Ramanan, D.,
  Doll{\'a}r, P., Zitnick, C.L.: Microsoft coco: Common objects in context.
\newblock In: European Conference on Computer Vision, pp. 740--755 (2014)

\bibitem{ASR}
Liu, B., Ding, Y., Jiao, J., Ji, X., Ye, Q.: Anti-aliasing semantic
  reconstruction for few-shot semantic segmentation.
\newblock In: Proceedings of the IEEE/CVF Conference on Computer Vision and
  Pattern Recognition, pp. 9747--9756 (2021)

\bibitem{liu2021cross}
Liu, W., Kong, X., Hung, T.Y., Lin, G.: Cross-image region mining with region
  prototypical network for weakly supervised segmentation.
\newblock IEEE Transactions on Multimedia  (2021)

\bibitem{liu2020guided}
Liu, W., Lin, G., Zhang, T., Liu, Z.: Guided co-segmentation network for fast
  video object segmentation.
\newblock IEEE Transactions on Circuits and Systems for Video Technology
  (2020)

\bibitem{QGNet}
Liu, W., Wu, Z., Ding, H., Liu, F., Lin, J., Lin, G.: Few-shot segmentation
  with global and local contrastive learning.
\newblock arXiv preprint arXiv:2108.05293  (2021)

\bibitem{liu2022long}
Liu, W., Wu, Z., Wang, Y., Ding, H., Liu, F., Lin, J., Lin, G.: Long-tailed
  recognition by learning from latent categories.
\newblock arXiv preprint arXiv:2206.01010  (2022)

\bibitem{liu2021few}
Liu, W., Zhang, C., Ding, H., Hung, T.Y., Lin, G.: Few-shot segmentation with
  optimal transport matching and message flow.
\newblock IEEE Transactions on Multimedia  (2021)

\bibitem{liu2020weakly}
Liu, W., Zhang, C., Lin, G., Hung, T.Y., Miao, C.: Weakly supervised
  segmentation with maximum bipartite graph matching.
\newblock In: Proceedings of the 28th ACM International Conference on
  Multimedia, pp. 2085--2094 (2020)

\bibitem{crnet}
Liu, W., Zhang, C., Lin, G., Liu, F.: Crnet: Cross-reference networks for
  few-shot segmentation.
\newblock In: Proceedings of the IEEE/CVF Conference on Computer Vision and
  Pattern Recognition, pp. 4165--4173 (2020)

\bibitem{ppnet}
Liu, Y., Zhang, X., Zhang, S., He, X.: Part-aware prototype network for
  few-shot semantic segmentation.
\newblock In: European Conference on Computer Vision, pp. 142--158. Springer
  (2020)

\bibitem{long2015fully}
Long, J., Shelhamer, E., Darrell, T.: Fully convolutional networks for semantic
  segmentation.
\newblock In: Proceedings of the IEEE conference on computer vision and pattern
  recognition, pp. 3431--3440 (2015)

\bibitem{citec}
Min, J., Kang, D., Cho, M.: Hypercorrelation squeeze for few-shot segmentation.
\newblock In: Proceedings of the IEEE/CVF International Conference on Computer
  Vision, pp. 6941--6952 (2021)

\bibitem{mukherjee2018object}
Mukherjee, P., Lall, B., Lattupally, S.: Object cosegmentation using deep
  siamese network.
\newblock arXiv preprint arXiv:1803.02555  (2018)

\bibitem{fast-weight}
Nguyen, K., Todorovic, S.: Feature weighting and boosting for few-shot
  segmentation.
\newblock In: Proceedings of the IEEE International Conference on Computer
  Vision, pp. 622--631 (2019)

\bibitem{peng2017large}
Peng, C., Zhang, X., Yu, G., Luo, G., Sun, J.: Large kernel matters--improve
  semantic segmentation by global convolutional network.
\newblock In: Proceedings of the IEEE conference on computer vision and pattern
  recognition, pp. 4353--4361 (2017)

\bibitem{rakelly2018conditional}
Rakelly, K., Shelhamer, E., Darrell, T., Efros, A., Levine, S.: Conditional
  networks for few-shot semantic segmentation.
\newblock In: International Conference on Learning Representations Workshop
  (2018)

\bibitem{ren2015faster}
Ren, S., He, K., Girshick, R., Sun, J.: Faster r-cnn: Towards real-time object
  detection with region proposal networks.
\newblock In: Advances in neural information processing systems, pp. 91--99
  (2015)

\bibitem{rother2006cosegmentation}
Rother, C., Minka, T., Blake, A., Kolmogorov, V.: Cosegmentation of image pairs
  by histogram matching-incorporating a global constraint into mrfs.
\newblock In: 2006 IEEE Computer Society Conference on Computer Vision and
  Pattern Recognition (Proceedings of the IEEE/CVF Conference on Computer
  Vision and Pattern Recognition'06), vol.~1, pp. 993--1000. IEEE (2006)

\bibitem{rubinstein2013unsupervised}
Rubinstein, M., Joulin, A., Kopf, J., Liu, C.: Unsupervised joint object
  discovery and segmentation in internet images.
\newblock In: Proceedings of the IEEE conference on computer vision and pattern
  recognition, pp. 1939--1946 (2013)

\bibitem{shaban2017one}
Shaban, A., Bansal, S., Liu, Z., Essa, I., Boots, B.: One-shot learning for
  semantic segmentation.
\newblock arXiv preprint arXiv:1709.03410  (2017)

\bibitem{siam2019adaptive}
Siam, M., Oreshkin, B.: Adaptive masked weight imprinting for few-shot
  segmentation.
\newblock arXiv preprint arXiv:1902.11123  (2019)

\bibitem{amp}
Siam, M., Oreshkin, B.N., Jagersand, M.: Amp: Adaptive masked proxies for
  few-shot segmentation.
\newblock In: Proceedings of the IEEE International Conference on Computer
  Vision, pp. 5249--5258 (2019)

\bibitem{snell2017prototypical}
Snell, J., Swersky, K., Zemel, R.: Prototypical networks for few-shot learning.
\newblock In: Advances in neural information processing systems (2017)

\bibitem{pfenet}
Tian, Z., Zhao, H., Shu, M., Yang, Z., Li, R., Jia, J.: Prior guided feature
  enrichment network for few-shot segmentation.
\newblock IEEE transactions on pattern analysis and machine intelligence
  (2020)

\bibitem{tip-co-seg2}
Tsai, C.C., Li, W., Hsu, K.J., Qian, X., Lin, Y.Y.: Image co-saliency detection
  and co-segmentation via progressive joint optimization.
\newblock IEEE Transactions on Image Processing \textbf{28}(1), 56--71 (2018)

\bibitem{vinyals2016matching}
Vinyals, O., Blundell, C., Lillicrap, T., Wierstra, D., et~al.: Matching
  networks for one shot learning.
\newblock In: Advances in neural information processing systems, pp. 3630--3638
  (2016)

\bibitem{danet}
Wang, H., Zhang, X., Hu, Y., Yang, Y., Cao, X., Zhen, X.: Few-shot semantic
  segmentation with democratic attention networks.
\newblock In: European Conference on Computer Vision, pp. 730--746. Springer
  (2020)

\bibitem{wang2019panet}
Wang, K., Liew, J.H., Zou, Y., Zhou, D., Feng, J.: Panet: Few-shot image
  semantic segmentation with prototype alignment.
\newblock In: Proceedings of the IEEE International Conference on Computer
  Vision, pp. 9197--9206 (2019)

\bibitem{sagnet}
Xie, G., Liu, J., Xiong, H., Shao, L.: Scale-aware graph neural network for
  few-shot semantic segmentation.
\newblock In: Proceedings of the IEEE/CVF Conference on Computer Vision and
  Pattern Recognition, pp. 5475--5484 (2021)

\bibitem{pmm}
Yang, B., Liu, C., Li, B., Jiao, J., Ye, Q.: Prototype mixture models for
  few-shot semantic segmentation.
\newblock arXiv preprint arXiv:2008.03898  (2020)

\bibitem{relation}
Yang, F.S.Y., Zhang, L., Xiang, T., Torr, P.H., Hospedales, T.M.: Learning to
  compare: Relation network for few-shot learning.
\newblock In: Proceedings of the IEEE/CVF Conference on Computer Vision and
  Pattern Recognition (2018)

\bibitem{cited}
Yang, L., Zhuo, W., Qi, L., Shi, Y., Gao, Y.: Mining latent classes for
  few-shot segmentation.
\newblock In: Proceedings of the IEEE/CVF International Conference on Computer
  Vision, pp. 8721--8730 (2021)

\bibitem{yosinski2015understanding}
Yosinski, J., Clune, J., Nguyen, A., Fuchs, T., Lipson, H.: Understanding
  neural networks through deep visualization.
\newblock arXiv preprint arXiv:1506.06579  (2015)

\bibitem{zhang2020deepemd}
Zhang, C., Cai, Y., Lin, G., Shen, C.: Deepemd: Few-shot image classification
  with differentiable earth mover's distance and structured classifiers.
\newblock In: Proceedings of the IEEE/CVF conference on computer vision and
  pattern recognition, pp. 12203--12213 (2020)

\bibitem{zhang2021meta}
Zhang, C., Ding, H., Lin, G., Li, R., Wang, C., Shen, C.: Meta navigator:
  Search for a good adaptation policy for few-shot learning.
\newblock In: Proceedings of the IEEE/CVF International Conference on Computer
  Vision, pp. 9435--9444 (2021)

\bibitem{zhangchi2}
Zhang, C., Lin, G., Liu, F., Guo, J., Wu, Q., Yao, R.: Pyramid graph networks
  with connection attentions for region-based one-shot semantic segmentation.
\newblock In: Proceedings of the IEEE International Conference on Computer
  Vision, pp. 9587--9595 (2019)

\bibitem{zhang2019canet}
Zhang, C., Lin, G., Liu, F., Yao, R., Shen, C.: Canet: Class-agnostic
  segmentation networks with iterative refinement and attentive few-shot
  learning.
\newblock In: Proceedings of the IEEE Conference on Computer Vision and Pattern
  Recognition, pp. 5217--5226 (2019)

\bibitem{zhang2020splitting}
Zhang, T., Lin, G., Liu, W., Cai, J., Kot, A.: Splitting vs. merging: Mining
  object regions with discrepancy and intersection loss for weakly supervised
  semantic segmentation.
\newblock In: European Conference on Computer Vision, pp. 663--679. Springer,
  Cham (2020)

\bibitem{zhang2018sg}
Zhang, X., Wei, Y., Yang, Y., Huang, T.: Sg-one: Similarity guidance network
  for one-shot semantic segmentation.
\newblock arXiv preprint arXiv:1810.09091  (2018)

\end{thebibliography}

\end{document}